\documentclass[a4paper,fleqn]{cas-sc}

\usepackage[numbers]{natbib}
\usepackage{algorithm}
\usepackage{algorithmic}
\usepackage{graphicx}
\usepackage{epstopdf}
\usepackage{subfigure}
\usepackage{soul}
\usepackage{multicol}
\usepackage{wrapfig}
\usepackage{booktabs}
\usepackage{adjustbox}
\usepackage{multirow}
\usepackage{bm}
\usepackage{amsmath, amsthm, amssymb}
\usepackage{pgfplots}
\usepackage{color}

\usepackage{hyperref}
\hypersetup{
colorlinks=true,
allcolors=DarkSlateGrey,
}

\pgfplotsset{compat=1.17}

\newcommand{\tabincell}[2]{\begin{tabular}{@{}#1@{}}#2\end{tabular}}
\renewcommand{\thefootnote}[1]{\textcolor{#1}{\arabic{footnote}}}
\renewcommand\thefootnote{\textcolor{red}{\arabic{footnote}}}

\def\tsc#1{\csdef{#1}{\textsc{\lowercase{#1}}\xspace}}
\tsc{WGM}
\tsc{QE}
\begin{document}
\let\WriteBookmarks\relax
\def\floatpagepagefraction{1}
\def\textpagefraction{.001}

\shorttitle{}    

\shortauthors{S. Tao, Q. Cao, H. Shen et al.}  

\title [mode = title]{Adversarial Camouflage for Node Injection Attack on Graphs}  

%\tnotemark[<tnote number>] 

%\tnotetext[<tnote number>]{<tnote text>} 


\author[1,3]{Shuchang Tao}[orcid=0000-0001-6113-6145]

%\cormark[<corr mark no>]

%\fnmark[<footnote mark no>]

\ead{taoshuchang18z@ict.ac.cn}

%\ead[url]{<URL>}

\credit{Adversarial attack, Graph neural networks, Data mining}

\author[1]{Qi Cao}
% Footnote of the second author

% Email id of the second author
\ead{caoqi@ict.ac.cn}


\credit{}
\cormark[1]

\author[1,3]{Huawei Shen}
% Footnote of the second author
%\fnmark[3]

% Email id of the second author
\ead{shenhuawei@ict.ac.cn}


\credit{}
\cormark[1]

\author[1,3]{Yunfan Wu}
% Footnote of the second author
%\fnmark[3]

% Email id of the second author
\ead{wuyunfan19b@ict.ac.cn}


\credit{}

\author[1,3]{Liang Hou}
% Footnote of the second author
%\fnmark[3]

% Email id of the second author
\ead{houliang17z@ict.ac.cn}


\credit{}

\author[1]{Fei Sun}
%% Footnote of the second author
%%\fnmark[3]
\ead{sunfei@ict.ac.cn}
%% URL of the second author
\credit{}

\author[2,3]{Xueqi Cheng}
% Footnote of the second author
%\fnmark[3]
\ead{cxq@ict.ac.cn}

\credit{}

\affiliation[1]{organization={Data Intelligence System Research Center, Institute of Computing Technology, Chinese Academy of Sciences},
            city={Beijing},
%          citysep={}, % Uncomment if no comma needed between city and postcode
            country={China}}
            
\affiliation[2]{organization={CAS Key Laboratory of Network Data Science and Technology, Institute of Computing Technology, Chinese Academy of Sciences},
            city={Beijing},
%          citysep={}, % Uncomment if no comma needed between city and postcode
            country={China}}
            
\affiliation[3]{organization={University of Chinese Academy of Sciences},
            city={Beijing},
            country={China}}

\cortext[1]{Corresponding author}



\begin{abstract}
Node injection attacks on Graph Neural Networks (GNNs) have received increasing attention recently, due to their ability to degrade GNN performance with high attack success rates. However, our study indicates  that these attacks often fail in practical scenarios, since defense/detection methods can easily identify and remove the injected nodes. To address this, we devote to \emph{camouflage node injection attack}, making injected nodes appear normal and imperceptible to defense/detection methods. Unfortunately, the non-Euclidean structure of graph data and  the lack of intuitive prior present great challenges to the formalization, implementation, and evaluation of camouflage. In this paper, we first propose and define camouflage as distribution similarity between ego networks of injected nodes and normal nodes. Then for implementation, we propose an \emph{adversarial \underline{CA}mouflage framework for \underline{N}ode injection \underline{A}ttack}, namely CANA, to improve attack performance under defense/detection methods in practical scenarios. A novel camouflage metric is further designed under the guide of distribution similarity. Extensive experiments demonstrate that CANA can significantly improve the attack performance under defense/detection methods with higher camouflage or imperceptibility. This work urges us to raise awareness of the security vulnerabilities of GNNs in practical applications.
\end{abstract}


%\item We propose a novel graph adversarial immunization problem.

%$
%Y \Perp D \mid Z_{c}
%$

% Keywords

\maketitle

% Main text
%\section{}\label{}

% Use the style of numbering in square brackets.
% If nothing is used, default style will be taken.
%\begin{enumerate}[a)]
%\item 
%\item 
%\item 
%\end{enumerate}  

%\begin{itemize}
%\item 
%\item 
%\item 
%\end{itemize}  

%\begin{description}
%\item[]
%\item[] 
%\item[] 
%\end{description}  

% Figure
%\begin{figure}[<options>]
%	\centering
%		\includegraphics[<options>]{}
%	  \caption{}\label{fig1}
%\end{figure}

%\begin{table}[<options>]
%\caption{}\label{tbl1}
%\begin{tabular*}{\tblwidth}{@{}LL@{}}
%\toprule
%  &  \\ % Table header row
%\midrule
% & \\
% & \\
% & \\
% & \\
%\bottomrule
%\end{tabular*}
%\end{table}

% Uncomment and use as the case may be
%\begin{theorem} 
%\end{theorem}

% Uncomment and use as the case may be
%\begin{lemma} 
%\end{lemma}


\section{Introduction}
\label{sec:intro}
Graph Neural Networks (GNNs) have achieved great success in various graph mining tasks, such as node classification~\cite{kipf2017semi,Xu2019HowPA}, graph classification~\cite{Xu2019HowPA}, cascade prediction~\cite{Cao2020PopularityPO}, and recommender systems~\cite{WangToseefurther2023,liao_sociallgn_2022}. 
Despite their success, GNNs have been proved to be vulnerable to graph adversarial attacks~\cite{Dai2018AdversarialAO,zugner2018adversarial},
i.e., small perturbations on graph data~\cite{zugner_adversarial_2019, zugner2018adversarial} can easily fool GNNs. 
These attacks can be categorized as graph modification attacks and node injection attacks~\cite{ZouTDGIA}. 
Specifically, graph modification attacks modify existing edges~\cite{zugner_adversarial_2019,zugner2018adversarial} or node attributes~\cite{zugner2018adversarial}.
 Node injection attacks focus on a more executable scenario, where attackers inject malicious nodes~\cite{Sun2020AdversarialAO,Wang2020ScalableAO,TaoGNIA,ZouTDGIA} rather than modifying existing node features or edges like graph modification attacks.
Node injection attacks have shown excellent attack performance~\cite{TaoGNIA,ZouTDGIA}, e.g., more than 90\% nodes can be successfully attacked even when only injecting one malicious node with one edge~\cite{TaoGNIA}. 
%The success of node injection attacks pose a serious threat to the security of GNNs in practical applications. 
%exposes the vulnerability of GNN, and has significant implications for the security of GNN. 

\begin{figure}
\centering
 \subfigure[G-NIA. Detect Acc=98\%]{
 \includegraphics[width=0.249\textwidth]{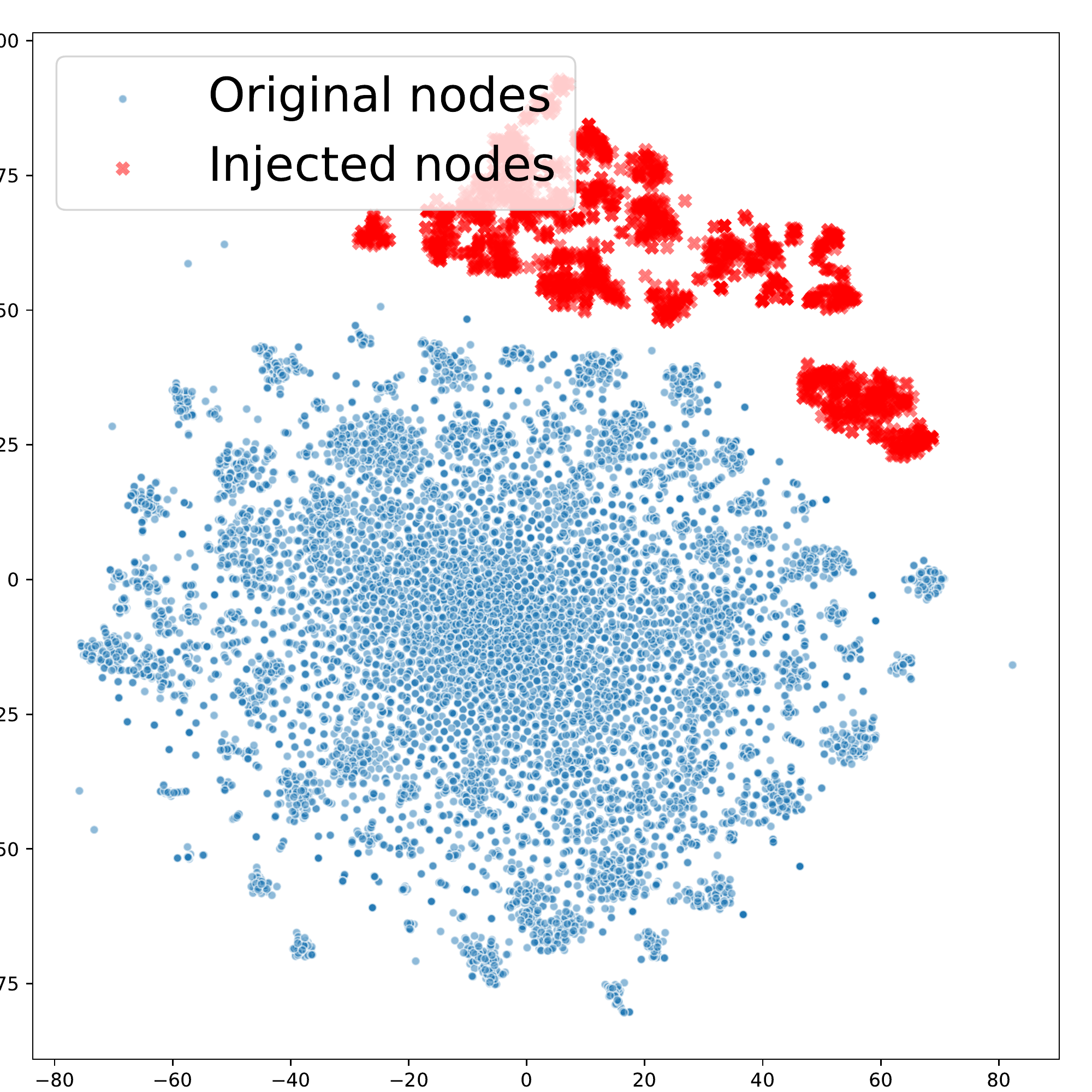}
 \label{subfig:feat_gnia}
 }\hspace{10mm}
 \subfigure[G-NIA+HAO. Detect Acc=91\%]{
 \includegraphics[width=0.25\textwidth]{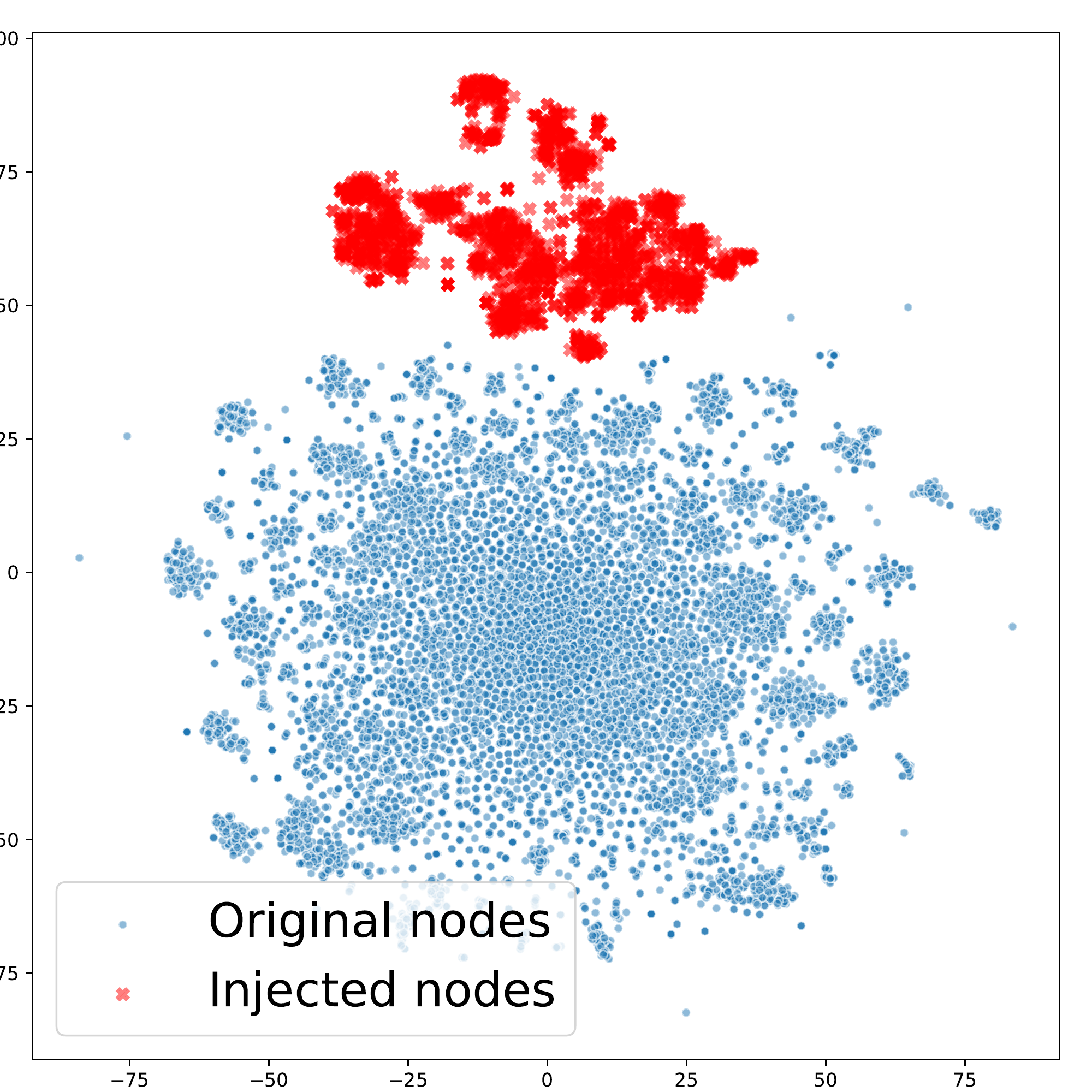}
 \label{subfig:feat_gnia}
 }\hspace{10mm}
 \subfigure[G-NIA+CANA. Detect Acc=36\%]{
 \includegraphics[width=0.25\textwidth]{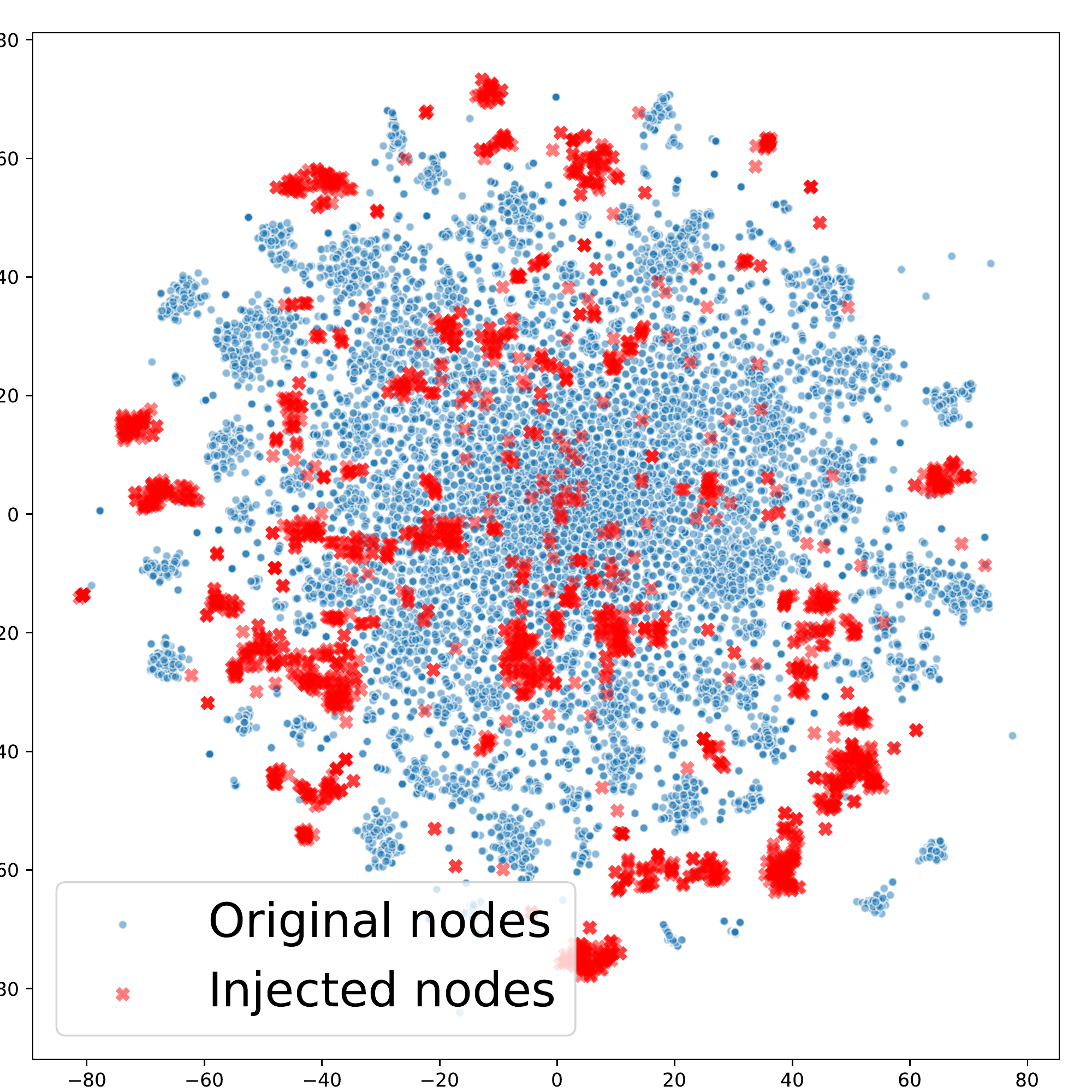}
 \label{subfig:feat_gnia+CANA}
 }\\
\subfigure[TDGIA. Detect Acc=64\%]{
\includegraphics[width=0.25\textwidth]{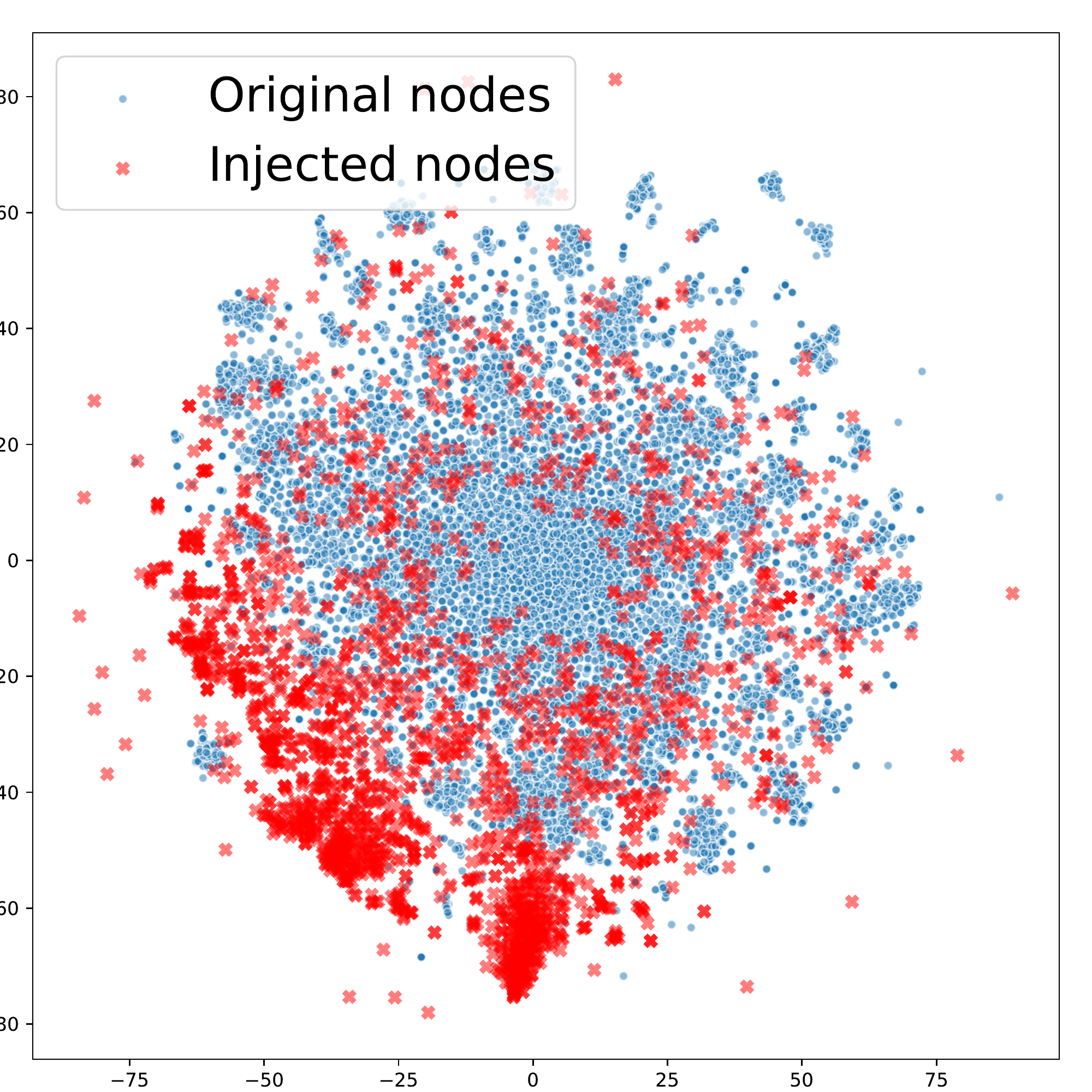}
\label{subfig:feat_tdgia}
%\end{minipage}%
}\hspace{10mm}
\subfigure[TDGIA+HAO. Detect Acc=92\%]{
\includegraphics[width=0.25\textwidth]{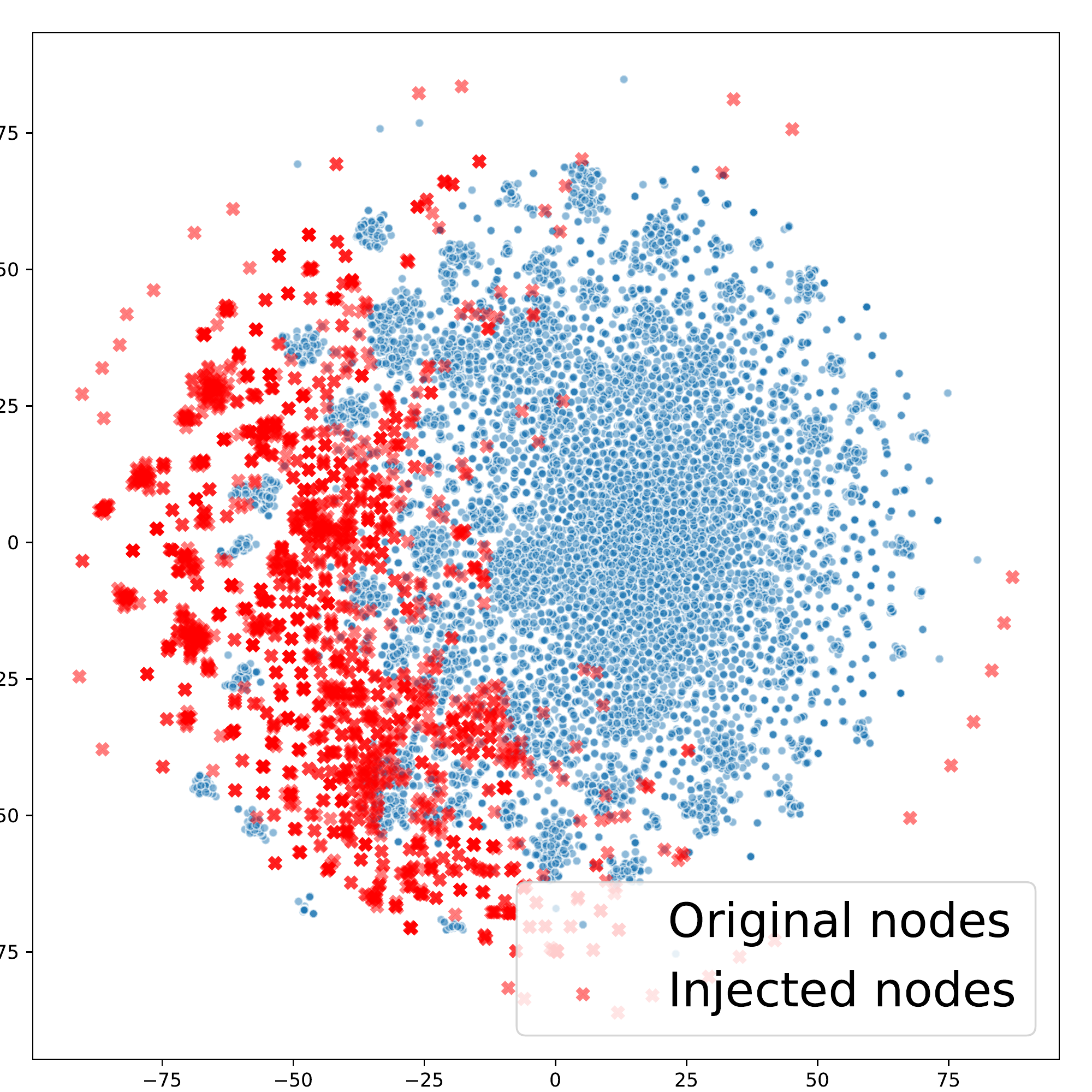}
\label{subfig:feat_tdgia}
%\end{minipage}%
}\hspace{10mm}
\subfigure[TDGIA+CANA. Detect Acc=38\%]{
\includegraphics[width=0.25\textwidth]{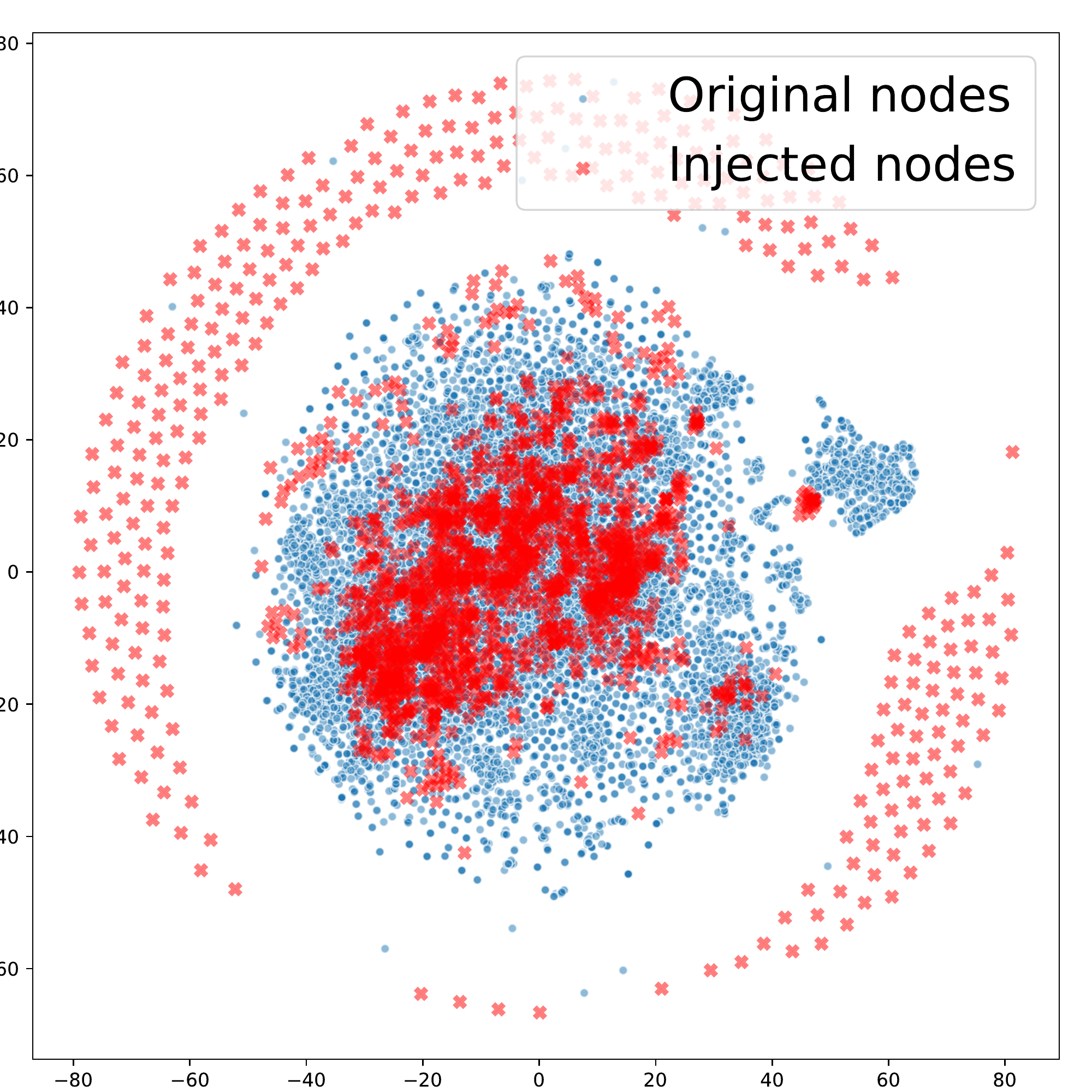}
\label{subfig:feat_tdgia+CANA}
}
\caption{T-SNE visualizations of attributes of injected nodes (red) and original normal nodes (blue). The detection accuracy of the detection methods on the nodes injected by each attack method is reported. Smaller means better imperceptibility.}
\label{fig:vis_feat}
%\vspace{-6pt}
\end{figure}

%
%\label{apd:detect}
%\begin{figure}
%\centering
%
%\caption{The detection and defense performance of five detection methods on TDGIA, TDGIA+HAO, and TDGIA+CANA: (a) Accuracy of detecting injected nodes. Smaller means better imperceptibility. (b) Misclassification rate of node classification task after detection filtering the injected nodes. Larger means better attack performance.}
%\label{fig:detect_tdgia}
%% \vspace{-8pt}
%\end{figure}
%

%Despite their success in no detection, we observe that the nodes injected by existing methods are easy to be distinguished from the original normal nodes, making the attacks fail.
Despite the success of node injection attacks in a simplified scenario without defense or detection, our study suggests that they can be prone to failure in practical situations. This is due to that defense and detection methods can easily distinguish and remove the injected malicious nodes from the original normal nodes.
Taking the ogbn-products dataset~\cite{Hu2020OGB} as an example, Figure~\ref{fig:vis_feat} shows the  attribute distribution of both injected nodes and original normal nodes for state-of-the-art node injection attack methods, i.e., G-NIA~\cite{TaoGNIA} and TDGIA~\cite{ZouTDGIA}, and heuristic imperceptible constraint HAO~\cite{chen2022understanding}.
Even with a simple t-SNE visualization method, we observe that the node attributes of injected nodes (red) look different from the normal ones (blue). 
%Such a clear distinction between injected nodes and normal nodes provides a loophole for defense or detection methods to identify and remove injected nodes, rendering attacks ineffective.
We also report the average detection accuracy of the state-of-the-art unsupervised anomaly detection methods~\cite{COPOD,PCA,AE,IForest,HBOS} for the injected nodes in Figure~\ref{fig:vis_feat}.
Specifically, the nodes injected by G-NIA are almost all detected (detection accuracy=98\%), which means almost all injected nodes can be removed, making the attacks fail\footnote{Further introduction on detection methods and detail analysis are shown in Section~\ref{sec:limit}.}.
Equipped with the imperceptible constraint HAO, the injected nodes of G-NIA+HAO become  more imperceptible, but the detection accuracy is still 91\%.
The defects weaken the effectiveness of such attacks in practical scenarios where defense/detection methods are commonly used.
Addressing this issue is critically important.
%Moreover, these attacks undermine the effectiveness of evaluating robustness, further compounding the gravity of the problem.
However, up to now, little attention has been paid to these defects of node injection attacks.

In this paper, we devote to \emph{camouflage node injection attack}, which makes injected nodes as the normal ones considering both structures and attributions to be imperceptible to defense or detection methods. 
Unfortunately, the non-Euclidean nature of graph data and the lack of intuitive prior present significant challenges to camouflage node injection attack from three aspects: 
\textbf{i) Definition of camouflage.} Since graph data are difficult to visualize and understand, it lacks intuitive priors to define the camouflage of injected nodes, which is the first important problem to be solved~\cite{Sun2018AdversarialAA}. 
\textbf{ii) Camouflage attack.} The challenge stems from how to achieve a camouflage node injection attack while maintaining the performance of existing node injection attack methods. \textbf{iii) Evaluation for camouflage.}  The semantic complexity of graph data presents challenges for evaluation, similar to those encountered in defining it. 
%Existing works~\cite{ZouTDGIA,TaoGNIA,chen2022understanding} mainly focus on adding constraints to develop imperceptible attack methods, rather than proposing a metric to evaluate imperceptibility or camouflage.
%Existing evaluation metrics for imperceptibility or camouflage are only focusing on specific heuristic aspects, such as limiting feature co-ocurence~\cite{zugner2018adversarial}, or relying on certain prior information such as smoothness~\cite{chen2022understanding}, fail to achieve a general evaluation metric.
%There is still a lack of a general metric to evaluate node camouflage. 

To address the \textbf{Challenge i},  we first propose and define the camouflage of injected nodes as the distribution similarity between the ego networks centered around the injected nodes and those centered around the normal nodes.
%This distribution similarity of ego networks considers both the network structures and node attributes. 
This definition, which utilizes the original normal nodes to guide the imperceptibility of injected nodes, is intuitive and easy to understand.
Our camouflage is more comprehensive and general compared to existing methods. Specifically, we integrate both network structure and node attributes into our camouflage definition, which is more comprehensive than the existing method~\cite{ZouTDGIA}  that only considers node attributes. 
Also, our proposed camouflage does not make any assumptions about the graph type. Thus, our camouflage is  more general compared with HAO~\cite{chen2022understanding} which is only applicable to homogeneous graphs, but not to other types of graphs such as heterophily graphs\footnote{In heterophily graphs~\cite{ZouGCGL23}, connected nodes may have different class labels and dissimilar features.}. 
%Our camouflage does not make any assumptions about the graph type and is thus more widely applicable.
%Specifically, fidelity means camouflaged injected nodes should be similar to the original ones, while diversity refers to that the injected nodes should not be the same as each other. 
To address the \textbf{Challenge ii}, we design an \emph{adversarial \underline{CA}mouflage framework for \underline{N}ode Injection \underline{A}ttack}, namely CANA, to improve the camouflage while ensuring the attack performance. 
Specifically, CANA is under a generative adversarial framework~\cite{goodfellow2014generative, MaoLTM019}, dedicated to generating malicious nodes that are difficult to differentiate from normal nodes by an adversarial camouflage discriminator.
Note that CANA is a general framework that can be attached to existing node injection attacks. Figure~\ref{subfig:feat_gnia+CANA} and Figure~\ref{subfig:feat_tdgia+CANA} demonstrate that the incorporation of our proposed CANA leads to a significant reduction in the distinction between injected nodes and the original normal nodes.
When tackling \textbf{Challenge iii}, measuring the similarity between distributions can be difficult due to the unknown probability densities of the distributions.
Inspired by Fr\'echet Inception Distance (FID)~\cite{Heus2017FID}, we propose the Graph Fr\'echet Distance (\textbf{GraphFD}) as a metric to compute the distance using the respective means and covariances of the ego network distributions of normal and injected nodes.
The proposed metric is further shown to be consistent with detection accuracy. 

% we further propose a novel metric, Graph Fr\'echet Distance (GraphFD), under the guide of distribution similarity. GraphFD computes the distance between the distributions of the ego networks of the injected nodes and the original normal nodes.

Extensive experiments are conducted on three kinds of network datasets across four node injection attack methods. When equipping existing node injection attack methods with our proposed CANA framework, the attack performance under defense and detection methods as well as node camouflage is significantly improved.

In a nutshell, our main contributions are as follows:
% \vspace{-2pt}
 \begin{itemize}
 	\item Our study reveals that existing node injection attacks are prone to failure in practical scenarios because defense or detection methods can easily distinguish and remove the injected malicious nodes from the original normal nodes.
 	\item We propose and formulate the camouflage of injected nodes, and design a general framework CANA for node injection attacks, which can be attached to existing methods to improve the camouflage and the attack performance in practical scenarios with defense/detection methods.
 	\item  We further propose a novel metric to evaluate camouflage under the guide of distribution similarity. Extensive experiments demonstrate that the integration of our CANA framework improves the attack performance  under defense/detection methods and camouflage level of node injection attacks.
 \end{itemize}

%$\bullet$ We find that the injected nodes by existing methods are easy to be identified by defense methods, limiting their attack performance in the real world.
%
%$\bullet$ We formulate the camouflage of injected nodes and propose an adversarial camouflage framework for node injection attacks, which can be attached to any existing method to improve their camouflage.
%
%$\bullet$ We further propose a novel metric to evaluate the imperceptibility. Extensive experiments demonstrate that equipped with our CANA, node injection attack methods improve attack performance on detection and defense methods, as well as imperceptibility on all the evaluation metrics.

\section{Related Work}
This section briefly reviews existing adversarial attack and defense methods for GNNs, as well as existing studies on the imperceptibility of adversarial attacks. 

\subsection{Adversarial Attacks for GNNs}
GNNs have achieved immense success in various graph mining tasks~\cite{zou_dgsln_2023, liao_sociallgn_2022, lee_hapgnn_2022,Huang2021sbgnn,Wu2022inmo}.
However, they are proven to be sensitive to adversarial attacks. Attackers can perturb both graph structure and node attributes to dramatically degrade the performance of GNNs~\cite{Jin2020AdversarialAA,zugner2018adversarial,Sun2018AdversarialAA, wu_parameter_2021}. Unfortunately, modifying existing edges~\cite{zugner_adversarial_2019} or node features of the original graph~\cite{zugner2018adversarial} requires high authority and makes the attack difficult to perform in practice~\cite{Sun2020AdversarialAO}.
Node injection attack~\cite{Sun2020AdversarialAO, Wang2020ScalableAO} focuses on a more practical scenario, which only injects some malicious nodes without modifying original node features or edges.
Specifically, AFGSM~\cite{Wang2020ScalableAO} provides an approximate closed-form attack solution for attacking a specific GNN,  i.e., SGC~\cite{pmlr-v97-wu19e}. 
NIPA~\cite{Sun2020AdversarialAO} uses hierarchical reinforcement learning to sequentially generate the labels and edges of the malicious nodes.
Unfortunately, NIPA suffers from high computational costs and is not scalable for large-scale datasets~\cite{ZouTDGIA}.
%Pioneering researches who focus on node injection poisoning attacks have some problematic disadvantages which cannot be overlooked in an evasion attack scenario. 
%Node Injection Poisoning Attack (NIPA) uses a hierarchical Q-learning network to sequentially generate the labels and edges of the malicious nodes~\cite{Sun2020AdversarialAO}.
%Nevertheless, NIPA fails to generate the attributes of injected nodes, which are extremely important in evasion attacks, resulting in its poor performance.
To offer a more flexible framework, TDGIA~\cite{ZouTDGIA} heuristically selects the defective edges for injecting nodes and adopts smooth optimization to generate features for injected nodes. Recently, G-NIA~\cite{TaoGNIA} studies an extremely limited scenario of single node injection attack, showing excellent attack performance when only injecting one malicious node. These attacks attract increasing research attention~\cite{tao2023immune, TaoGNIA, ZouTDGIA, Sun2018AdversarialAA} due to their high success rate. 
%In the following, we mainly explore these node injection attacks with strong attack performance, such as TDGIA and G-NIA.

While existing node injection attacks demonstrate high success rates, our study reveals these attacks to be prone to failure under defense/detection methods, which can easily identify the injected malicious nodes.
In real-world scenarios where defense and detection methods are commonly employed, these easy-to-detect attacks will be ineffective.
%Moreover, these attacks undermine the effectiveness of evaluating robustness, further compounding the gravity of the problem.
%In addition, AFGSM cannot deal with the continuous attributed graphs which are common in the real world.
%Such approximation also leads to an underutilization of node attributes and structure.  

\subsection{Imperceptibility of Adversarial Attacks}
Adversarial attacks should be imperceptible to defense/detection methods~\cite{LiuZJLL22},. In computer vision, the perturbations on digital adversarial examples are generally regularized by $\ell_1$ norm, $\ell_2$ norm, or $\ell_\infty$ norm, while the perturbations in physical-world adversarial examples require a more practical constraint, e.g., appear to be legitimate to human observers. 
To accomplish this imperceptibility goal, multiple techniques have been explored, including the use of natural styles to disguise physical-world adversarial samples~\cite{LiuZJLL22,ChenZXSS20,ZhangTSZZL23}, rendering them more convincing to human observers

As for the imperceptibility of adversarial attacks on graphs, existing researchers simply perform imperceptible attacks by adopting heuristic constraints on specific aspects. For example, researchers either limit the budget of modified edges/nodes~\cite{Sun2020AdversarialAO,Wang2020ScalableAO, TaoGNIA,ZouTDGIA}, or heuristically constrains the feature co-occurrence~\cite{zugner2018adversarial}, or injected feature range~\cite{ZouTDGIA}. For the recent node injection attacks, HAO~\cite{chen2022understanding} tries to improve the unnoticeability of node injection attacks by introducing smoothness or so-called homophily constraint. However, this approach remains a one-sided heuristic and requires the graph to have a high degree of smoothness.

\subsection{Defense/detection methods for GNNs}
In order to deal with attacks, lots of methods have been proposed~\cite{Sun2018AdversarialAA, Jin2020AdversarialAA, tao2021advimmune, tao2023immune, wu_ergcn_2022, tao2023idea} to detect or defense under graph adversarial attacks. 

\textbf{Anomaly Detections.}
%The detection methods~\cite{ioannidis2019graphsac,xu2018characterizing} generally protect the GNNs by exploring the difference between perturbed and clean nodes/edges, removing malicious edges before applying GNNs. 
Anomaly detection methods have achieved great success in many real-world applications such as fraud detection and cyber intrusion detection~\cite{HBOS, PCA,AE}. They aim to find instances that are deviated from normal ones~\cite{IForest, COPOD}, which are appropriate for detecting the injected nodes.
They are also used to detect adversarial attacks~\cite{DetectAdv}.
Anomaly detection methods can be categorized into five types~\cite{ADBench}, namely probabilistic, linear model, proximity-based, outlier ensemble, and neural network methods.
We select state-of-the-art methods for each, specifically COPOD~\cite{COPOD} for probabilistic methods, PCA~\cite{PCA} for linear model methods, HBOS~\cite{HBOS} for proximity-based methods, IForest~\cite{IForest} for outlier ensemble methods, and AE~\cite{AE} for neural network methods. 
These methods are used to detect the injected nodes in this paper.

\textbf{Defense methods.}
The defense methods can be mainly categorized into~\cite{Jin2020AdversarialAA}: adversarial training and attention mechanism. 
Specifically, adversarial training~\cite{kong2020flag,li2022spectral} adopts a min-max optimization style, iteratively generating perturbations that maximize the loss and updating GNNs parameters that minimize the loss. 
The attention mechanism aims to train a robust GNN model by penalizing the model’s weights on adversarial edges or nodes~\cite{Jin2020AdversarialAA, ZhangGNNGuard2020, jin2021node}.
% To defend against poisoning attacks, graph purification~\cite{Entezari2020AllYN,kong2020flag} is proposed, which first purifies the perturbed graph and then trains the GNNs on the purified graph.

With the development of adversarial learning, more and more powerful defense methods are continuously proposed, just like an arms race between attackers and defenders. As a result, how to design attack methods that are imperceptible to detection/defense methods is one of the key points of adversarial attacks.

To sum up, there is still a lack of general definitions, effective methods, as well as evaluation metrics for the imperceptibility or camouflage of node injection attacks.
Hence, it urges us to study the camouflage on graphs.

% In computer vision adversarial learning, extensive researches focus on imperceptible attack based on generative adversarial networks (GANs)~\cite{goodfellow2014generative}.
% AdvGAN~\cite{xiao2018generating} first proposed to use the generator of GAN to generate adversarial samples and encourage adversarial samples to be as similar as possible to real samples through the discriminator.
% AdvGAN++~\cite{jandial2019advgan++} improved the attack performance against defense models of AdvGAN by introducing hidden features in a classifier as the input of the GAN to produce adversarial samples.
% AdvCGAN~\cite{wang2021advcgan} further utilized conditional GANs~\cite{mirza2014conditional} to elastically produce more realistic adversarial examples with any arbitrarily-assigned attack label and achieve higher attack accuracy, especially in targeted attack.
% Natural GAN~\cite{zhao2018generating} searched for representation of adversarial samples in the hidden feature space of the low-dimensional manifold, which makes the generated disturbance of adversarial samples more targeted and natural.
% % RobGAN~\cite{Liu_2019_CVPR},
% AdvCam~\cite{Duan_2020_CVPR} elaborately made and disguised adversarial samples of physical-world into natural styles by transferring perturbations into "hidden" on-target object or off-target background, making them reasonable to human observers.
% Advfaces~\cite{deb2020advfaces} inherited the framework of AdvGAN and introduced human identity matching information to generate adversarial face images.

\begin{figure}
%\begin{wrapfigure}{r}{0.5\textwidth}
\centering
\subfigure[G-NIA: Detection accuracy  $\downarrow$]{
\includegraphics[width=0.232\textwidth]{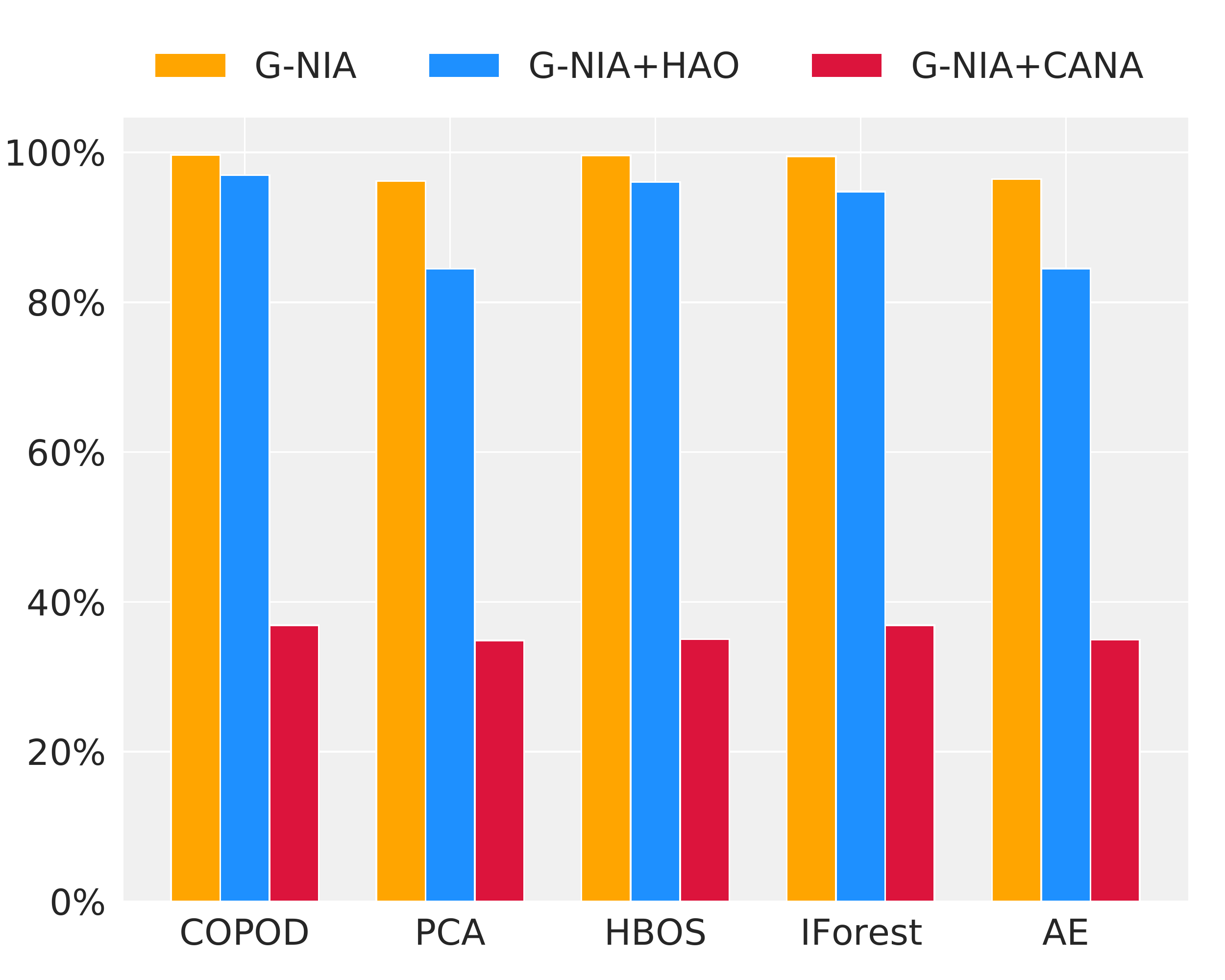}
}
\subfigure[G-NIA: Misclassification rate   $\uparrow$]{
\includegraphics[width=0.232\textwidth]{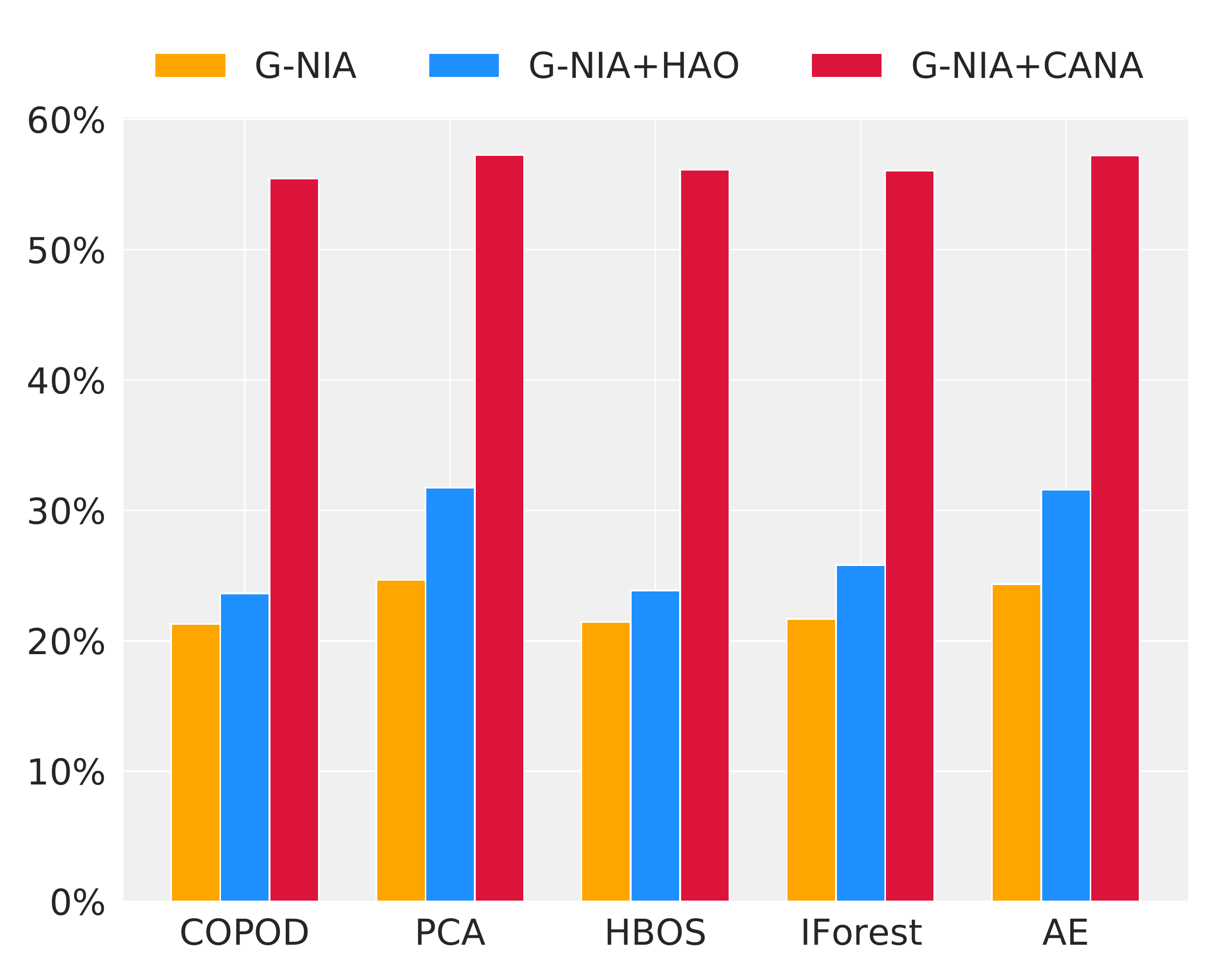}
}
\subfigure[TDGIA: Detection accuracy  $\downarrow$]{
\includegraphics[width=0.232\textwidth]{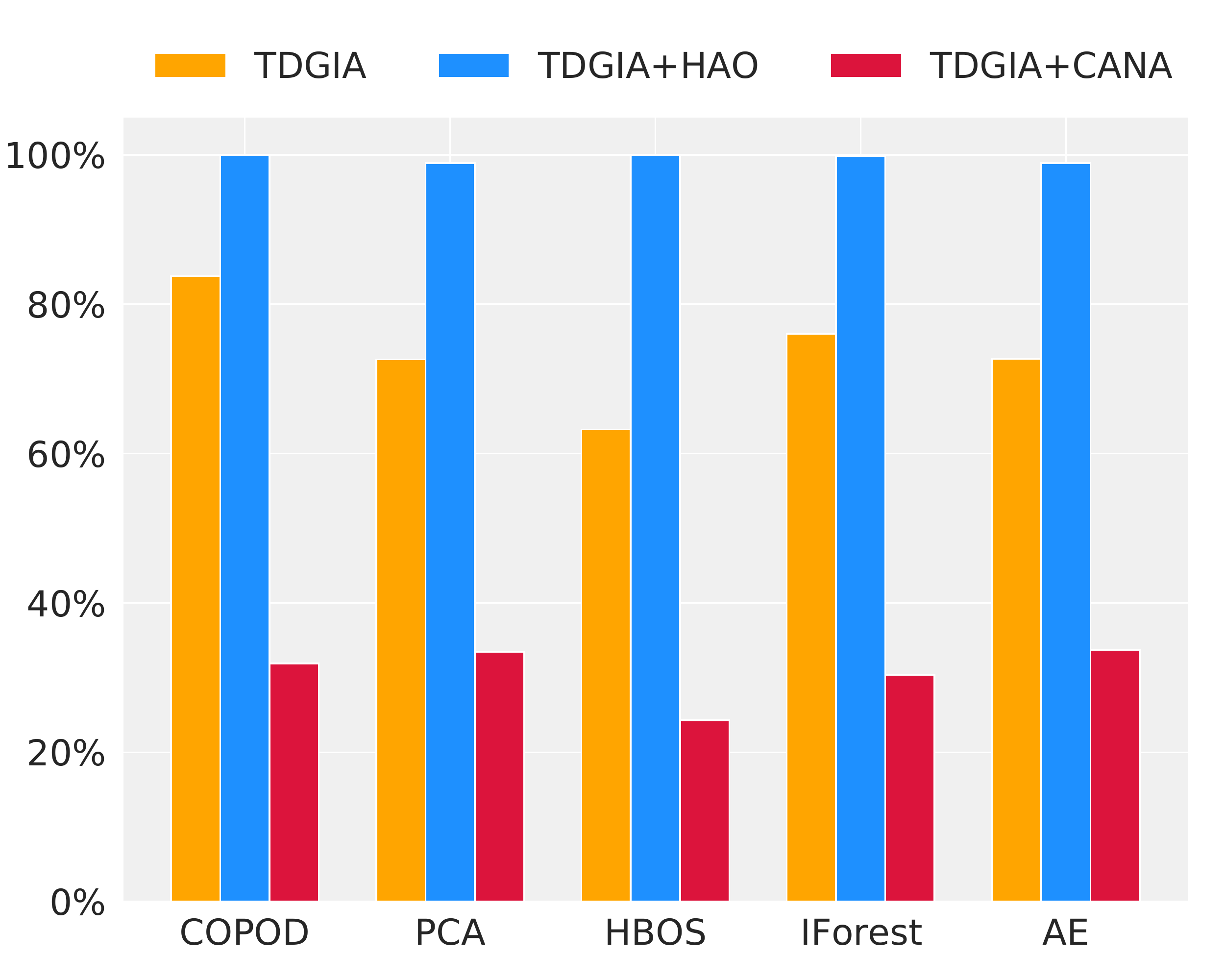}
}
\subfigure[TDGIA: Misclassification rate   $\uparrow$]{
\includegraphics[width=0.232\textwidth]{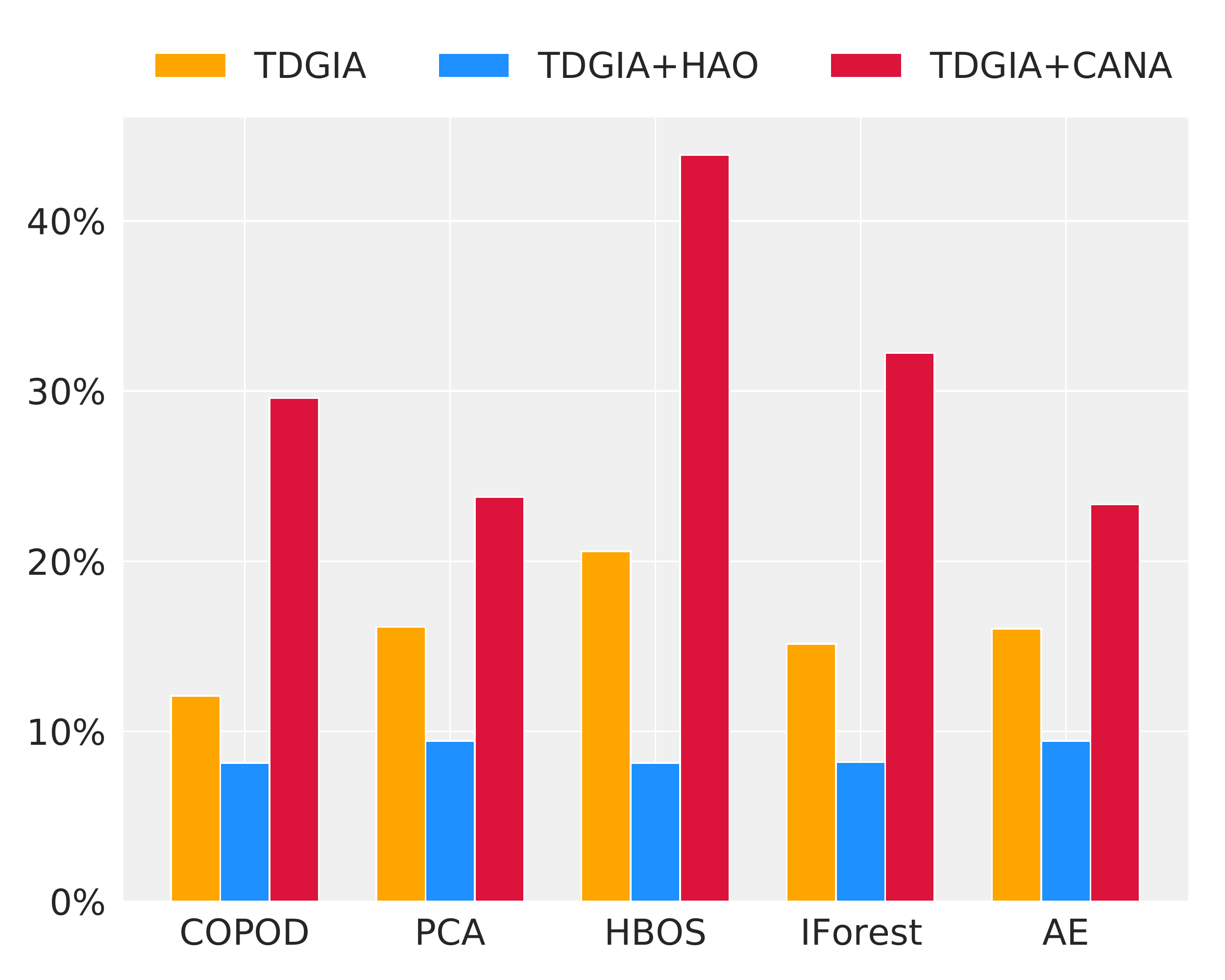}
}
\caption{The accuracy and defense performance of five detection methods. (a,c) Accuracy of detecting injected nodes. Smaller means better imperceptibility. (b,d) Misclassification rate of node classification task after detection filtering the injected nodes. Larger means better attack performance.}
\label{fig:detect}
%\vspace{-8pt}
%\end{wrapfigure}
\end{figure}

\section{Preliminaries}

In this section, we first introduce node injection attacks. Then we analyze the limitations of existing attacks.

\subsection{Node Injection Attack}
\textbf{Node classification.}
We use the widely adopted node classification task as the downstream task.
Let $G=(V,E,X)$ be an attributed graph, where $V=\{1,2,\dots,n\}$ is the set of $n$ nodes, $E\subseteq V\times V$ is the set of edges, and $X\in\mathbb{R}^{n\times d}$ is the attribute matrix of nodes with $d$-dimensional attributes.
%Given an  attributed graph $G=(A,X)$, $A=\{a_{vu}|v,u\in V\}$ is  the adjacency matrix and node attributes $X=\{x_v|v\in V\}$ where $V$ denotes the node set.
Given a class set $K$ of node labels, the task aims to predict the label of each node by learning a classifier $f_\theta$, i.e., $\boldsymbol{Z} = f_{\theta}(G) $, where $\boldsymbol{Z} \in \mathbb{R}^{n \times k}$ and $k=|K|$ is the number of classes.
For example, GNN is a commonly used classifier $f_{\theta}(\cdot)$ containing feature transformation and aggregation operations, which benefits many graph mining tasks.

\textbf{Node injection attack.}
The goal of node injection attacks is to degrade the performance of GNN by injecting malicious nodes rather than modifying original node features or edges.
 Specifically, the attacker $\mathcal{G}$ aims to mislead the prediction of a GNN model  $f_{\theta}$ on the target nodes $V_\text{tar}$ by injecting malicious nodes to obtain a perturbed graph.
Formally, the objective function of node injection attack is as follows~\cite{TaoGNIA}:
\begin{equation}
\begin{aligned}
&\max_{\mathcal{G}} \sum_{t\in V_\text{tar}} \mathbb{1}\left( \arg\max f_{\theta^{*}} \left(\mathcal{G}(G) \right)_t \neq y_t\right)  \\
&\mathcal{G}(G)=( V\cup \hat{V}, E\cup \hat{E} ,X \mathbin\Vert \hat{X}) \\
&s.t.\quad \theta^{*}=\arg \min _{\theta} \mathcal{L}_{\operatorname{train}}\left(f_{\theta}(\tilde{G}),y\right),
\label{eq:ogj} 
\end{aligned}
%\vspace{-1pt}
\end{equation}
%where $y_t$ is the ground truth of the target node $t$, $\mathcal{G}(G)$ is the perturbed graph obtained by the node injection attack method $\mathcal{G}$, $\mathbb{1}(\cdot)$ is a metric of whether the attack can successfully mislead the model prediction, $\arg \max f_{\theta^{*}} \left(\mathcal{G}(G) \right)_t$ is the predicted class of $f_{\theta^{*}}$ on node $t$ in the perturbed graph $\mathcal{G}(G)$.
where $y_t$ is the ground truth label of the target node $t$, $\mathcal{G}(G)$ represents the perturbed graph obtained after applying the node injection attack method $\mathcal{G}$ to the original graph $G$, $\mathbb{1}(\cdot)$ is a binary metric that indicates whether the attack is successful in misleading the model prediction, and $\arg \max f_{\theta^{*}} \left(\mathcal{G}(G) \right)_t$ represents the predicted class of the target node $t$ in the perturbed graph $\mathcal{G}(G)$ using the model $f_{\theta^{*}}$.
Based on the original graph $G=(V,E,X)$, the perturbed graph $\mathcal{G}(G)$ additionally includes the injected nodes $\hat{V}$, injected edges $\hat{E}$, and the corresponding node attributes $\hat{X}$.
$f_{\theta^{*}}(\tilde{G})$ is the victim GNN model trained on original graph $\tilde{G}=G$ or the perturbed graph $\tilde{G}=\mathcal{G}(G)$.

\subsection{Limitation of Existing Methods}
\label{sec:limit}
%Based on the objective described above, a successful attack on a graph neural network (GNN) would result in a difference between the predicted label of target nodes before and after the attack, assuming that these nodes were correctly classified prior to the attack. 
%This difference indicates that the GNN has detected the influence of the injected nodes.
%According to the above objective, a successful attack indicates that the predicted labels of target nodes before and after the attack should be different for GNN, assuming that these nodes are correctly classified without attacking. This means that GNN perceives the influence of the injected nodes.
Based on the objective described above, the injected nodes can be quite different from the surrounding normal nodes and easy to be perceived, which is demonstrated in our empirical study.
As illustrated in Figure~\ref{fig:vis_feat}, we find that the nodes injected by existing methods are perceptible, as they are different from the original normal nodes with a simple t-SNE visualization.
Anomaly detection methods can identify the outliers effectively, and show great success in real-world applications.
We adopt the state-of-the-art unsupervised anomaly detection methods (COPOD~\cite{COPOD}, PCA~\cite{PCA}, HBOS~\cite{HBOS}, IForest~\cite{IForest}, and AE~\cite{AE}) to detect injected nodes.
% We find detection methods can easily distinguish the injected nodes. 
Figure~\ref{fig:detect} (a) depicts the accuracy of detecting nodes injected by the state-of-the-art attack G-NIA, and imperceptible constraint HAO.
Almost all of the nodes injected by G-NIA can be detected by detection methods. G-NIA+HAO slightly improves the imperceptibility but still results in 91\% of the injected nodes being detected.
After removing the detected nodes, we input the obtained new graph into GNN and evaluate the misclassification rate. Figure~\ref{fig:detect} (b) illustrates the poor attack performance of G-NIA, i.e., 20\% of misclassification rate, under detection methods, which is a significant drop from the previous misclassification rates of around 99\% when detection methods were not applied.
We also show that our method (CANA) evades detection methods and achieves a high attack success rate. The implementations are elaborated in Section~\ref{sec:CANA}.
Additionally, we conduct experiments for another state-of-the-art node injection attack TDGIA and observe similar problems, as shown in Figure~\ref{fig:detect} (c) and (d). 
%Detection models are common in the real world. Such ease-to-detection of existing node injection attacks will reduce their performance in real scenarios. They also may not evaluate the robustness of real scenarios cannot be well evaluated
%The clear distinction between injected nodes and original nodes increases the risk of the injected nodes being identified, limiting the attack performance of existing methods in the real world.
The results demonstrate that existing node injection attacks are easy-to-detect and will fail in a practical scenario since detection models are commonly deployed in the real world.
%What's worse, this makes it difficult for existing attacks to evaluate the robustness in the practical scenario.

% However, up to now, there is still a lack of a clear definition of imperceptibility.

\section{Camouflage Node Injection attack}
In this section, we propose and formulate the camouflage of injected nodes from the distribution similarity of ego networks.
We also propose an adversarial camouflage framework for node injection attacks (CANA) to improve camouflage while ensuring attack performance for existing attacks.
\subsection{Definition of Camouflage}
\begin{figure*}[t]
\centering
	\includegraphics[width=\textwidth]{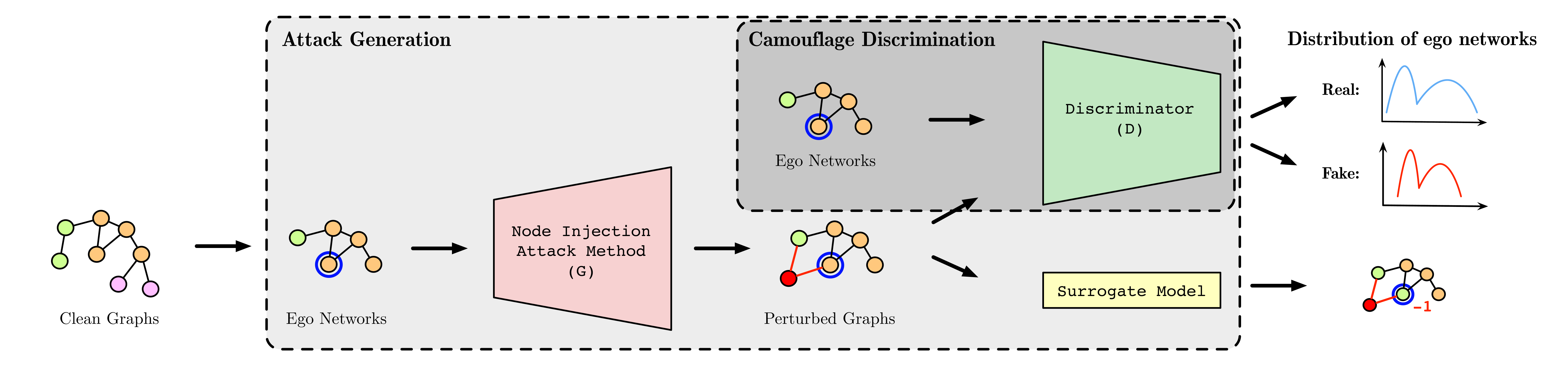}
	\caption{The overall architecture of CANA. 
	CANA adopts an adversarial paradigm, where the attack method (generator) aims to improve the camouflage while maintaining the attack performance. Discriminator aims to distinguish the distributions of ego networks between the injected nodes and the normal ones, providing informative guidance for the attacker.
	}
	\label{fig:frame}
%	\vspace{-6pt}
\end{figure*}

We are dedicated to \emph{camouflage node injection attack}. However, camouflage node injection attack is quite difficult, since the non-Euclidean nature of graph data makes it difficult to visualize and understand. 
To solve this issue, our intuition is that camouflaged injected nodes should be similar to the original nodes. Thus, an ideal state of camouflaged injected nodes is to follow the distribution of the original normal nodes.  
To capture both node attributes and network structure, we utilize the distribution of ego network~\cite{Freeman82Ego} which consists of a focal node (ego) and its $L$-hop neighbor nodes~\cite{wu2022handling}. We denote the $L$-hop neighbors of node $v$ as $N$, and the ego network as $G_v$, which consists of nodes $V_v=N_v$, edges $E_v=\{e_{uw}|u,w \in N_v\}$, and node attributes $X_v=\{x_u|u\in N_v\}$. In this paper, we adopt the two-hop ego network $G_v=(V_v, E_v, X_v)$. Examples of ego networks can be found in Figure~\ref{fig:frame}.
%We devote to \emph{camouflage node injection attack}, i.e., camouflaging injected malicious nodes (structure/attributes) as the normal ones that appear legitimate/imperceptible to detection or defense methods. 
%Unfortunately, defining the camouflage on graphs is not easy. 
%Due to the non-Euclidean nature of graph data, it is difficult to visualize and understand, resulting in little intuitive prior.
%G_v = (V_v=N_v, E_v=\{e_{uw}|u,w \in N_v\}, X_v=\{x_u|u\in N_v\}),

We propose to define the camouflage of injected nodes as the distribution similarity between the ego networks centering around the injected nodes and the ego networks centering around the normal nodes. 
In other words, the ego network distribution of the camouflaged injected nodes is similar to that of the original normal nodes, considering both the surrounding network structure and node attributes.
Mathematically, we express the camouflage as:
\begin{equation}
\begin{aligned}
P_{v \sim V}\left(G_{v}\right) \sim P_{\hat{v} \sim \hat{V}}\left(\mathcal{G}(G)_{\hat{v}}\right).
\end{aligned}
\label{eq:goal}
\end{equation}
Here, $G_{v}$ denotes the ego network of the normal node $v$ in the original graph $G$, $\mathcal{G}(G)_{\hat{v}}$ refers to the ego network of the injected node $\hat{v}$ in the perturbed graph $\mathcal{G}(G)$. $P_{V}\left(G_{v}\right)$ represents the distribution of ego networks centering around the original nodes, $P_{\hat{V}}\left(\mathcal{G}(G)_{\hat{v}}\right)$ indicates the distribution of ego networks centering around the injected nodes in the perturbed graph. 
To represent the distribution of ego networks, we encode each ego network as a vector and assemble these vectors as columns in a matrix. This matrix encapsulates the distribution of ego networks in the dataset.
A detailed explanation of ego network representation can be found in Section~\ref{sec:CANA}.
Note that the $\sim$ denotes the distribution similarity.  Ideally, camouflaged injected nodes satisfy $P_{V}\left(G_{v}\right) = P_{\hat{V}}\left(\mathcal{G}(G)_{\hat{v}}\right)$.
We formally establish the objective of camouflage using the Jensen–Shannon divergence as follows:
\begin{equation}
\begin{aligned}
\small &\min_{\mathcal{G}(G)} D_{\text{JS}}\bigl(P_{V}(G_{v}) \,\Vert\, P_{\hat{V}}(\mathcal{G}(G)_{\hat{v}})\bigl).
\end{aligned}
\label{eq:JS}
\end{equation}
So far, we establish the definition and objective of camouflage and  utilize the original normal nodes to guide the imperceptibility of injected nodes, providing a good guideline for solving the limitations of existing methods in Section~\ref{sec:limit}.
 Our camouflage surpasses existing constraints because it is not restricted to specific aspects, such as attribute similarity~\cite{ZouTDGIA} or smoothness~\cite{chen2022understanding}. 
 For instance, in contrast to the smoothness constraint that necessitates a high degree of local smoothness, our camouflage is predicated on distribution similarity and guided by the original normal nodes. This allows for the learning of an appropriate degree of smoothness without requiring any a priori assumptions about the graph. Consequently, our camouflage is more versatile in nature.
Moreover, it is reasonable to surmise that, in accordance with our camouflage definition, injected nodes can elude detection by prevailing methods. Detection mechanisms generally target nodes ~\cite{COPOD,PCA,AE,ADBench} that display atypical behavior or possess uncommon attributes or structures. Our camouflage aims to ensure that the distribution of injected nodes resembles that of the original normal nodes, thereby making the behavior of injected nodes more akin to normal nodes and facilitating evasion from existing detection methods.

%\begin{prop}
%\label{prop:definition}
%The optimal solution of Eq.~\ref{eq:JS} ,i.e., $D_{\text{JS}}(P_{V}(G_{v}) \| P_{\hat{V}}(\mathcal{G}(G)_{\hat{v}}))=0$, can achieve the camouflage injection nodes which will not be detected by the distance-based detection mechanisms. 
%\end{prop}
%
%\begin{proof}
%The optimal solution to Eq.~\ref{eq:JS} consists of the injected nodes and edges, whose ego networks are distributed in the same way as the original normal nodes, and are denoted as $(\hat{V}, \hat{E}, \hat{X})$.
%We assume $(\hat{V}, \hat{E}, \hat{X})$ can be detected by detection/defense methods, and reach a contradiction.
%% 数学形式化，按照什么原则的detection
%We discuss defense methods and detection methods separately.
%For the detection methods, the nodes that are detected are the ones exhibiting different behaviors from the majority of nodes, such as having abnormal node attributes $\hat{X}$ or structures $\hat{E}$.
%However, the solution  Eq.~\ref{eq:JS} 
%For the defense methods, 
%%The solution of Eq.~\ref{eq:JS} is the injected nodes whose ego networks follow the same distribution as the original normal nodes.
%JS divergence Assuming 
%\end{proof}

% 

\subsection{Adversarial Camouflage Framework for Node Injection Attack}
\label{sec:CANA}
%Howerver, since the meaning of camouflage is extensible, it is rather impossible for us to directly optimize all of them.
In order to incorporate the camouflage into node injection attacks, we design and propose \emph{adversarial \underline{CA}mouflage framework for \underline{N}ode Injection \underline{A}ttack}, namely CANA.
CANA is under a generative adversarial framework~\cite{goodfellow2014generative, MaoLTM019}, aiming to preserve the distribution similarity of ego networks.
Note that CANA is a general framework that can be equipped on existing node injection attacks, making them more camouflage or imperceptible.

Figure~\ref{fig:frame} illustrates the overall architecture of our proposed CANA.
CANA consists of two key components: node injection attack method $\mathcal{G}$ and adversarial discriminator $\mathcal{D}$.
The discriminator $\mathcal{D}$ learns the distinction between the distribution of the ego networks of injected nodes (fake) and normal nodes (real), providing camouflage guidance for attack method $\mathcal{G}$. 
%Meanwhile, the attacker generates perturbed graphs by having D inject nodes and edges into the original graph and computes three losses related to attack performance and camouflage for optimization.
Meanwhile, the attacker $\mathcal{G}$ injects nodes and edges into the original graph to generate perturbed graphs, as well as computes two kinds of losses involving attack performance and camouflage for optimization. 
For attack performance, the attacker learns to attack successfully under the guidance of the surrogate model.
For camouflage, the attacker optimizes the injected nodes with the guidance of $\mathcal{D}$ to make their distribution resemble that of the original nodes.
%To avoid mode collapse of injected nodes, diversity loss is further incorporated to maximize the distance between the ego networks of two injected nodes.
Next, we introduce the fine details of each component of our CANA.

\subsubsection{Camouflage Discrimination}
Camouflage is defined by the distribution of ego networks, thus we employ a discriminator to identify whether the ego network is real or fake and provide the camouflage information to guide the attacker.
%Discriminator aims to distinguish the ego network of injected nodes from that of original nodes.
The loss of camouflage discriminator is:
\begin{equation}
%\mathcal{L}_{\mathcal{D}}^\text{GAN} =\quad & \mathbb{E}_{x \sim P_{\text {data }}} \log \mathcal{D}\left(x_{\text{ego}}\right)\\
%&+ \mathbb{E}_{x^{\prime} \sim \mathcal{G}\left(t\right), t\sim P_\text{tar}} \log \left(1-\mathcal{D}\left(x^{\prime}_{\text{ego}}\right)\right).
\mathcal{L}_{\mathcal{D}}^\text{GAN} = -\mathbb{E}_{v \sim V} \log \mathcal{D}\left(G_{v}\right)
- \mathbb{E}_{\hat{v} \sim \hat{V}} \log \left(1-\mathcal{D}\left(\mathcal{G}(G)_{\hat{v}}\right)\right).
\label{eq:loss_d}
\end{equation}
Note that the discriminator needs to distinguish the real/fake label of ego networks, which is essentially a graph classification task.
We adopt Graph Isomorphism Networks (GIN)~\cite{Xu2019HowPA} as the discriminator, due to its provably powerful discriminative ability on the graph classification task.
By minimizing $\mathcal{L}_{\mathcal{D}}^\text{GAN}$, the discriminator learns the distribution similarity of ego networks of injected nodes and original normal nodes, to better assist the attacker.

% \begin{table*}[t]
% \caption{Statistics of the datasets}
% \label{tab:dataset}
% \begin{tabular}{c|cccccc|ccc}
% \toprule
% Dataset     &  $N_{LCC}$ &  $E_{LCC}$ & Classes & \tabincell{c}{Feature \\dimension} & \tabincell{c}{Average \\degree} & Feature range       & \tabincell{c}{Injected \\feature range} & \tabincell{c}{Injected \\node budget} & \tabincell{c}{Injected \\edge budget} \\
% \midrule
% ogbn-products & 10,494  & 38,872    & 35      & 100               & 3.7            & {[}-74.70,152.71{]} & {[}-20, 20{]}          & 2,099                 & 3                    \\
% reddit      & 10,004  & 73,512    & 41      & 602               & 7.35           & {[}-22.89, 80.85{]} & {[}-20, 20{]}          & 2,001                 & 7                    \\
% ogbn-arxiv    & 169,343 & 2,484,941  & 39      & 128               & 14.67          & {[}-1.39, 1.64{]}   & {[}-1, 1{]}            & 33,869                & 14      \\            
% \bottomrule      
% \end{tabular}
% \end{table*}

\subsubsection{Attack Generation}
The goal of the attack method $\mathcal{G}$ is to attack successfully while preserving camouflage.
%\textbf{imperceptibility Preservation.}
To ensure camouflage,
we encourage the attacker to generate malicious nodes that are difficult to be distinguished by discriminator $\mathcal{D}$.
Formally, the adversarial camouflage loss is: 
\begin{equation}
\begin{aligned}
%\mathcal{L}_{\mathcal{G}}^\text{GAN} &=-\mathbb{E}_{x^{\prime} \sim \mathcal{G}\left(t\right), t\sim P_\text{tar}} \log \mathcal{D}\left(x^{\prime}\right).
\mathcal{L}_{\mathcal{G}}^\text{GAN} &= - \mathbb{E}_{\hat{v} \sim \hat{V}} \log \mathcal{D}\left(\mathcal{G}(G)_{\hat{v}}\right).
\label{eq:loss_fid}
\end{aligned}
\end{equation}
According to the guide of $\mathcal{D}$, the attacker $\mathcal{G}$ learns to generate injected nodes distributed similarly to normal ones.
The main advantage of this loss is that it helps capture the underlying distribution of the original nodes, making it suitable for our definition of camouflage.

%\textbf{Diversity Preservation.}
To avoid mode collapse (injected nodes are too similar to each other to ensure camouflage well), we adopt diversity sensitive loss to maximize the distance between each pair of injected nodes~\cite{GuoHZXBD20,MaoLTM019}. The diversity sensitive loss is defined as:
\begin{equation}
%\begin{aligned}
%\mathcal{L}_{\mathcal{G}}^\text{div}=-\mathbb{E}_{u^{\prime}\sim \mathcal{G}\left(u\right), v^{\prime}\sim \mathcal{G}\left(v\right),u,v\sim P_\text{tar}}
%\sum_{l} d\left(\mathcal{R}^{l}\left(u^{\prime}_\text{ego}\right),\mathcal{R}^{l}\left(v^{\prime}_\text{ego}\right)\right),
\mathcal{L}_{\mathcal{G}}^\text{div}=-\mathbb{E}_{\hat{u}\sim \hat{V},\hat{v}\sim \hat{V}}
\sum_{l} d\left(\mathcal{R}^{l}\left(\mathcal{G}(G)_{\hat{u}}\right),\mathcal{R}^{l}\left(\mathcal{G}(G)_{\hat{v}}\right)\right),
\label{eq:loss_div}
%\end{aligned}
\end{equation}
where $d(\cdot)$ is the euclidean distance.
Note that the diversity sensitive loss is defined on the deep features, which is the output of representation model $\mathcal{R}$. $\mathcal{R}^{l}(\cdot)$ is the representation on $l^{th}$ layer of the ego network of an injected node. 
To better encode local graph patterns, we employ a pre-trained GIN as the representation model $\mathcal{R}$.
This loss encourages the attacker to generate diverse injected nodes for better camouflage.

For attack performance, we apply the attack loss following~\cite{TaoGNIA}: 
\begin{equation}
\mathcal{L}_{\mathcal{G}}^\text{atk}=\mathbb{E}_{t\sim V_{\text{tar}}}  \bigl[ \mathcal{F}(\mathcal{G}(G))_{t, y_t}-\max _{j \neq y_t} \mathcal{F}(\mathcal{G}(G))_{t, j}  \bigl],
\label{eq:loss_atk}
\end{equation}
where $t$ is sample from target nodes, $V_{\text{tar}}$, $\mathcal{F}(\cdot)$ is the surrogate model, $ \mathcal{F}(\mathcal{G}(G))_{t, y_t}$ is the output probability of $t$ on its ground truth label $y_t$ from $\mathcal{F}$,
$j$ is the class that is most likely to be misclassified among other classes.
This attack loss has been demonstrated to provide the best attack performance compared with other kinds of losses~\cite{Carlini2016TowardsET}.
Note that our CANA is a general framework, and $\mathcal{L}_{\mathcal{G}}^\text{atk}$ can be replaced by any node injection attack loss.

\begin{algorithm}[t]
\caption{Training Process for CANA Framework}
\begin{algorithmic}[1]
\label{alg:A}
\REQUIRE original graph $G=(V,E,X)$, pre-trained representation model $\mathcal{R}$, pre-trained surrogate model $f$
\ENSURE camouflage attack method $\mathcal{G}$
\FOR{number of training iterations}
  \FOR{number of discriminator training steps}
    \STATE{Sample minibatch of targets $V_{\text{tar}}$ from original normal nodes $V$}
   	\STATE{Generate $\mathcal{G}(G)=(V\cup \hat{V} ,E\cup \hat{E} ,X\mathbin\Vert \hat{X})$ by $\mathcal{G}$ attacking $V_{\text{tar}}$}
   	\STATE{Sample minibatch of clean nodes $v$ from $V$, and take the ego networks $G_{v}$}
   	\STATE{Sample minibatch of injected nodes $\hat{v}$ from $\hat{V}$,  and take the ego networks $\mathcal{G}(G)_{\hat{v}}$}
   	\STATE{Compute discriminator loss $\mathcal{L}_{\mathcal{D}}^\text{GAN}$ by Eq.~\ref{eq:loss_d} }
   	\STATE{Compute gradient of $\mathcal{D}$ and update $\mathcal{D}$}
  \ENDFOR
  \STATE{Sample minibatch of targets $V_{\text{tar}}$ from original normal nodes $V$}
  \STATE{Generate $\mathcal{G}(G)=(V\cup \hat{V} ,E\cup \hat{E} ,X\mathbin\Vert \hat{X})$ by $\mathcal{G}$ attacking $V_{\text{tar}}$}
  \STATE{Sample minibatch of injected nodes $\hat{v}$ from $\hat{V}$,  and take the ego networks $\mathcal{G}(G)_{\hat{v}}$}
  \STATE{Compute adversarial camouflage loss $\mathcal{L}_{\mathcal{G}}^\text{GAN}$ by Eq.~\ref{eq:loss_fid} }
  \STATE{Sample $(u,v)$ in $\hat{V}$, compute diversity sensitive loss $\mathcal{L}_{\mathcal{G}}^\text{div}$ by Eq.~\ref{eq:loss_div}}
  \STATE{Compute attack loss $\mathcal{L}_{\mathcal{G}}^\text{atk}$ by Eq.~\ref{eq:loss_atk}}
  \STATE{Compute overall loss $\mathcal{L}_{\mathcal{G}}$ by Eq.~\ref{eq:loss}}
  \STATE{Compute gradient of $\mathcal{G}$ and update $\mathcal{G}$}
  \ENDFOR
\end{algorithmic}
% \vspace{-3pt}
\end{algorithm}

\subsubsection{Overall Objective}

The full objective function of our proposed CANA can be summarized as follows:
\begin{equation}
\begin{aligned}
\min_{\mathcal{D}} \mathcal{L}_{\mathcal{D}} &= \mathcal{L}_{\mathcal{D}}^\text{GAN}, \\
\min_{\mathcal{G}} \mathcal{L}_{\mathcal{G}} &= \mathcal{L}_{\mathcal{G}}^\text{atk} + \alpha \mathcal{L}_{\mathcal{G}}^\text{GAN} + \beta\mathcal{L}_{\mathcal{G}}^\text{div},
\label{eq:loss}
\end{aligned}
\end{equation}
where $\alpha$ and $\beta$ are the weights to control the importance of camouflage and diversity, respectively.
The discriminator $\mathcal{D}$ and generator $\mathcal{G}$ are updated alternatively until convergence.
%Specifically, we update the discriminator $k$ steps and then $1$ generator step to ensure convergence.
%We provide the algorithm in Appendix~\ref{apd:alg}.

\subsection{Algorithm}
\label{apd:alg}
We describe the training procedure of CANA in Algorithm~\ref{alg:A}.
For each iteration, we first train the discriminator for several steps, and then train the attack generator once, following ~\cite{goodfellow2014generative}.
The complexity of CANA depends on the underlying basic attacks. Let the computational complexity of a basic attack be denoted by $\mathcal{O}(T)$, the complexity after incorporating the CANA framework becomes $\mathcal{O}((h+1) \cdot T)$. Here, $h$ represents the number of discriminator training steps, which is usually a small constant. 
%The difference is that CANA generates perturbed graphs attacking target nodes, instead of generating samples from noises,
%In addition, CANA optimizes the diversity loss to avoid mode collapse.

\section{Experiments}
In this section, we equip state-of-the-art node injection attacks with our CANA framework on three benchmark datasets and compare the attack performance  under representative detection and defense methods.
Further analysis of camouflage, case study, visualization, ablation study, and hyper-parameter analysis are also provided.

\subsection{Experimental Settings}
\subsubsection{Datasets}

We take the commonly-used node classification as the target task, following~\cite{zugner2018adversarial,zugner_adversarial_2019,TaoGNIA}. 
To illustrate the wide adaptability of CANA, we conduct experiments on three kinds of real-world datasets: a social network reddit~\cite{TaoGNIA} gathered from the Reddit website, an Amazon product co-purchasing network ogbn-products~\cite{Hu2020OGB}, and a citation network obgn-arxiv~\cite{Hu2020OGB} containing Computer Science (CS) arXiv papers.
\begin{itemize}
	\item \emph{reddit}. Each node represents a post, with word vectors as attributes and community as the label, while each edge represents the post-to-post relationship.
	\item \emph{ogbn-products}. A node represents a product sold on Amazon with the word vectors of descriptions as attributes and the product category as the label, and edges between two products indicate that they are purchased together.
	\item \emph{ogbn-arxiv}. A node denotes a CS arXiv paper with attributes indicating the average word embeddings of its title and abstract. An edge means the citation relationship and the node label is the primary category of each paper.  
	\end{itemize}

Experiments are conducted on the largest connected components (LCC) for all three network datasets, following~\cite{zugner2018adversarial,zugner_adversarial_2019,TaoGNIA}. 
The statistics of each dataset are summarized in Table~\ref{tab:dataset}.
%The statistics of each dataset are summarized in Table~\ref{tab:dataset}.

%Note that we limit the range of injected node features, and the budget of injected nodes and edges to ensure that the injected node satisfies the basic camouflage constraints.
%The statistics of each dataset are summarized in Table~\ref{tab:dataset}.

\begin{table}[t]
\centering
\caption{Statistics of the datasets. ${LCC}$ denotes the number of nodes in the largest connected components (LCC).}
\label{tab:dataset}
\resizebox{\textwidth}{!}{
\begin{tabular}{c|rrrrrr|rrr}
\toprule
Dataset     &  $N_{LCC}$ &  $E_{LCC}$ & Classes  & \tabincell{c}{Degree} & \tabincell{c}{Feature \\dimension}  & Feature range       & \tabincell{c}{Injected \\feature range} & \tabincell{c}{Injected \\node budget} & \tabincell{c}{Injected \\edge budget} \\
\midrule
ogbn-products & 10,494  & 38,872    & 35                    & 3.70         & 100                         & {[}-74.70,152.71{]} & {[}-20, 20{]}          & 2,099                 & 3                                        \\
reddit      & 10,004  & 73,512    & 41                    & 7.35                     & 602                          & {[}-22.89, 80.85{]} & {[}-20, 20{]}          & 2,001                 & 7                  \\
ogbn-arxiv    & 169,343 & 2,484,941  & 39                     & 14.67           & 128              & {[}-1.39, 1.64{]}   & {[}-1, 1{]}            & 33,869                & 14       \\            
\bottomrule      
\end{tabular}}
\end{table}

\subsubsection{Attack Methods}
Since CANA serves as a general camouflage framework for node injection attacks, we instantiate CANA with several state-of-the-art node injection attack methods to verify the generality and effectiveness of CANA. 
We employ the PGD~\cite{Madry2017TowardsDL}, TDGIA~\cite{ZouTDGIA} and G-NIA~\cite{TaoGNIA} as the basic attack methods due to their high attack success rate and excellent  scalability.
%~\footnote{ \textcolor{blue}{AFGSM~\cite{Wang2020ScalableAO} and NIPA~\cite{Sun2020AdversarialAO} are excluded due to the following reasons: AFGSM is only for attacking a specific GNN (SGC), NIPA faces high computational costs, making it not scalable for large-scale datasets.}}.
We also adopt heuristic imperceptible constraint HAO~\cite{chen2022understanding} as our strong baseline. Note that all attacks are evasion attacks following~\cite{ZouTDGIA,TaoGNIA}. 
As for the attack scenario, we utilize the commonly-used black-box attack, which is considered as the most practical scenario in real-world applications 
\footnote{The black-box attack scenario assumes that the attacker has no access to the model structure, training parameters, or defense methods (if defense methods are added) of the victim model and can only interact with the model through input and output.}.
\begin{itemize}
	\item \emph{Basic attack methods}.
\begin{itemize}
\vspace{-5pt}
	\item \textbf{PGD}~\cite{Madry2017TowardsDL} is a widely adopted attack, using projected gradient descent to generate malicious nodes.
	\item \textbf{TDGIA}~\cite{ZouTDGIA} adopts topologically defective edge selection and smooth feature optimization to inject malicious nodes.
	\item \textbf{G-NIA}~\cite{TaoGNIA}  models the injection attacking process via a parametric model to ensure the attack efficiency and attack performance.
\end{itemize}
%We adopt PGD to substitute basic gradient-based AFGSM
	\item \emph{Imperceptible constraint}.
\begin{itemize}
\vspace{-5pt}
\item \textbf{HAO}~\cite{chen2022understanding} is our most related work, which tries to promote the unnoticeability of node injection attacks using homophily constraint. 
\end{itemize}
%However, such homophily constraint is only a heuristic one-sided embodiment of the distinction between the injected nodes and the original ones, which is still hard to comprehensively improve the imperceptibility or camouflage of the injected nodes.
\end{itemize}

\subsubsection{Detection and Defense Methods}
To better demonstrate the attack performance of node injection methods and the improvement of CANA, we evaluate attacks on SOTA anomaly detection and defense methods, which form a practical adversarial environment.
\begin{itemize}
	\item \emph{Detection methods} identify outlier items that deviate from the majority of data distribution~\cite{ADBench} and are used in many applications such as fraud detection.  Anomaly detections can be divided into five types~\cite{ADBench}, including probabilistic, linear model, proximity-based, and neural network methods. For each category, we select the SOTA method, i.e., COPOD~\cite{COPOD}, PCA~\cite{PCA}, HBOS~\cite{HBOS}, IForest~\cite{IForest}, and AE~\cite{AE}.
	\item \emph{Defense methods} defend adversarial attacks via adversarial training, or attention mechanism~\cite{Sun2018AdversarialAA, Jin2020AdversarialAA}.
For each category, we adopt the most representative methods.
For adversarial training, we leverage FLAG~\cite{kong2020flag} which is efficient and effective.
%For detection, we select the best anomaly detection methods. 
For the attention mechanism, we adopt the powerful model GNNGuard~\cite{ZhangGNNGuard2020}.
\end{itemize}

%$\bullet$ \emph{Adversarial training} is a widely used countermeasure for adversarial attacks, the main idea of which is to inject adversarial examples into the training set such that the trained model can correctly classify the future adversarial examples~\cite{Jin2020AdversarialAA}. Here, following~\cite{chen2022understanding}, we adopt an efficient adversarial training~\cite{10.1145/3097983.3098061} for attention mechanism based graph neural networks~\cite{Zhu2019RobustGC}, i.e., FLAG~\cite{kong2020flag}, to defense adversarial attacks, referred to as \textbf{Adv}.

%The hyperparameters settings are in Appendix~\ref{apd:hyp}.

\subsubsection{Hyperparameters Settings}
\label{apd:hyp}
We implement all the attack methods with our CANA framework.
For all attack methods, we limit the range of injected node features, as well as the budget of injected nodes and edges to ensure that the perturbation is subtle, as shown in Table~\ref{tab:dataset}. 
In each dataset, we randomly choose $20\%$ nodes as the target nodes and  inject the same number of malicious nodes to perform the attack. 
For parametric G-NIA, we first train it on a part of non-target nodes and then infer the injected nodes for the target nodes based on the trained model. Other methods are directly optimized and obtain the injected nodes for the target nodes.
The attack method $\mathcal {G}$ and discriminator $\mathcal{D}$ are trained alternatively following GAN~\cite{goodfellow2014generative}, i.e., we update the discriminator $h$ steps and then optimize the generator $1$ step to ensure the convergence. 
%Note that, we stop updating discriminator $\mathcal{D}$ when it could accurately classify 80\% of nodes, even if $\mathcal{D}$ has been optimized fewer than $h$ steps.
The surrogate model and representation model are pre-trained on original normal node labels.
%Adam optimizer is employed to train all the parameters. 
%The overall training procedure is finished when both of misclassification rate and camouflaged metric GraphFD (Section~\ref{sec:metric}) are converged.
We tune the hyper-parameters based on the validation performance within the following range: $\alpha$  over $\{0.1, 0.5, 1.0, 5.0\}$, $\beta$ over $\{10^{-3}, 10^{-2}, 10^{-1}, 1\}$, and $h$ over $\{1, 4, 10\}$, and learning rate over $\{10^{-3}, 10^{-2}, 10^{-1}\}$.
\subsection{Metrics for Camouflage Evaluation}
\label{sec:metric}
Existing works~\cite{ZouTDGIA,TaoGNIA,chen2022understanding} mainly focus on adding constraints to introduce imperceptible attack methods, rather than proposing a metric to evaluate imperceptibility or camouflage.
Based on these constraints, we develop several metrics to evaluate the camouflage of injected nodes.
Specifically, TDGIA~\cite{ZouTDGIA} limits the range of attributes based on attribute similarity. Inspired by this, we pose the Closest Attribute Distance (\textbf{CAD}), which measures the average Euclidean distance between the attribute and its most similar attribute of original normal nodes. Formally, $\text{CAD}_i = \min_{j\in V} d(X_{i}, X_j)$.
HAO~\cite{chen2022understanding} constrains the homophily of injected nodes, i.e., the similarity between two connected nodes, also known as smoothness. 
Building on this, we introduce Smoothness (\textbf{Smooth}) metric, which is the average distance between a node's attribute and those of its neighbors. Formally, $\text{Smooth}_i = \frac{1}{|N_i|} \sum_{j\in N_i} d(X_i, X_j)$.
However, these metrics only focus on specific heuristic aspects. There is still a lack of a general metric to evaluate camouflage from the perspective of distribution similarity.

\begin{figure}
%\begin{wrapfigure}{r}{0.5\textwidth}
\centering
\subfigure[GraphFD and the perceptible level of injected nodes.]{
\includegraphics[width=0.42\textwidth]{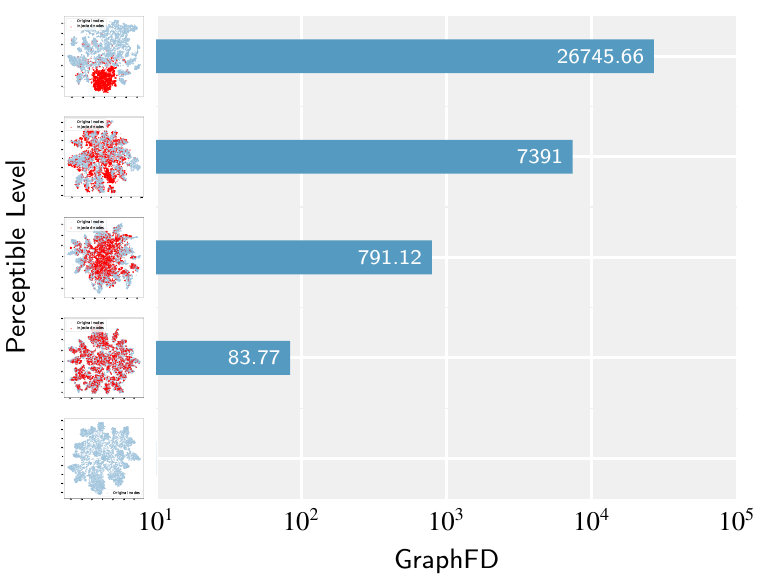}
\label{subfig:gfd_percep}
}\hspace{5pt}
\subfigure[GraphFD and the accuracy of detecting injected nodes.]{
\includegraphics[width=0.42\textwidth]{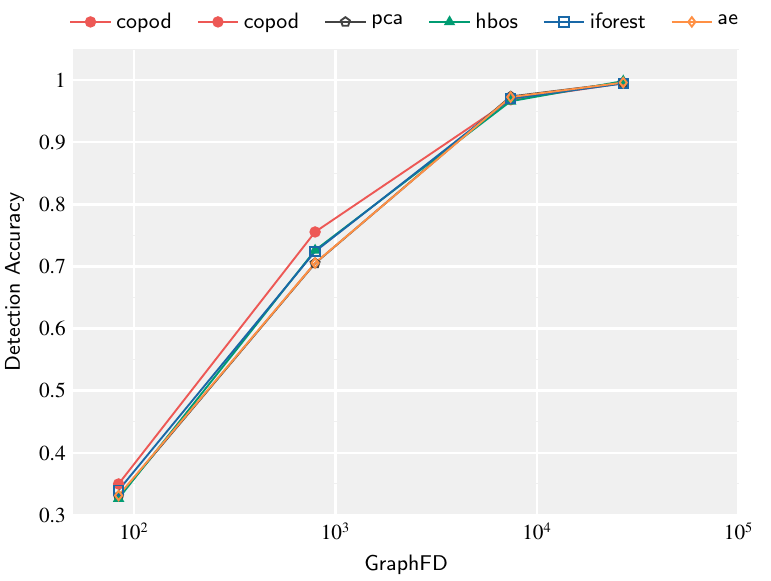}
\label{subfig:gfd_detect}
}
\caption{The rationality of GraphFD. (a) The relation between GraphFD and the perceptible level of injected nodes based on the t-SNE visualization. (b) The relation between GraphFD and the accuracy of detecting injected nodes.}
\label{fig:gfd}
%\vspace{-8pt}
%\end{wrapfigure}
\end{figure}

\begin{table*}[t]
\caption{Misclassification rate (\%) of node injection attack across detection and defense methods on ogbn-products, reddit, and ogbn-arxiv. The relative increases in Average and Median misclassification rates compared with basic attacks are written in parentheses.}
\label{tab:atk}
\resizebox{\textwidth}{!}{
\begin{tabular}{l l ccccc cc l l}
\toprule
\multicolumn{1}{l}{}            &           & \multicolumn{5}{c}{Detection methods}                                              & \multicolumn{2}{c}{Defense methods} & \multicolumn{1}{c}{\multirow{2}{*}{Average}}   & \multicolumn{1}{c}{\multirow{2}{*}{Median} }   \\
\cmidrule(lr){3-7} \cmidrule(lr){8-9}
\multicolumn{1}{l}{}            & \textbf{} & COPOD          & PCA            & HBOS           & IForest        & AE    & FLAG        & GNNGuard         &                            &                            \\
\midrule
\multirow{10}{*}{\tabincell{c}{ogbn-\\products}} & Clean     & 25.20          & 25.54          & 24.87          & 25.30          & 25.49          & 21.11            & 19.29            & 23.83                     & 25.20                     \\
 \cmidrule(lr){2-11}
                                & PGD       & 36.68          & 37.73          & 36.11          & 37.49          & 37.49          & 29.87            & 33.21            & 35.51                     & 36.68                     \\
                                & \quad +HAO   & 31.25 & 32.16 & 32.11 & 33.35   & 31.92       & 27.06        & 33.49                & 31.62                     & 32.11                    \\
                                & \quad +CANA     & \textbf{40.78} & \textbf{42.73} & \textbf{43.59} & \textbf{44.97} & \textbf{42.78} & \textbf{31.92}   & \textbf{35.11}   & \textbf{40.27 (  $\uparrow$ 13.39)}  & \textbf{42.73 (  $\uparrow$ 16.49)}  \\
                                \cmidrule(lr){2-11}
                                & TDGIA     & 37.26          & 46.21          & 42.21          & 44.07          & 46.02          & \textbf{33.35}   & \textbf{39.21}   & 41.19                     & 42.21                     \\
                                & \quad +HAO      & 24.92          & 31.30          & 25.01          & 25.96          & 31.30          & 25.58            & 33.68            & 27.31                     & 27.35                     \\
                                & \quad +CANA     & \textbf{45.97} & \textbf{49.36} & \textbf{45.55} & \textbf{48.79} & \textbf{49.21} & 27.49            & 30.63            & \textbf{42.43 (  $\uparrow$ 3.01)}   & \textbf{45.97 (  $\uparrow$ 8.92)}   \\                       
                                \cmidrule(lr){2-11}
                                & G-NIA     & 21.30          & 24.68          & 21.44          & 21.68          & 24.34          & 32.92            & 32.25            & 25.51                     & 24.34                     \\
                                & \quad +HAO      & 23.63          & 31.73          & 23.87          & 25.82          & 31.59          & 21.06            & 21.92            & 25.03                     & 25.54                     \\
                                & \quad +CANA     & \textbf{55.45} & \textbf{57.27} & \textbf{56.12} & \textbf{56.07} & \textbf{57.22} & \textbf{40.11}   & \textbf{34.21}   & \textbf{50.92 (  $\uparrow$ 99.58)}   & \textbf{56.07 (  $\uparrow$ 130.33)} \\                                
\midrule
\multirow{10}{*}{reddit}        & Clean     & 9.25           & 9.25           & 9.25           & 9.15           & 9.25           & 8.05             & 17.59            & 10.25                     & 9.25                      \\
 \cmidrule(lr){2-11}
                                & PGD       & 9.70           & 14.94          & 10.29          & 11.44          & 14.59          & 17.99            & 24.79            & 14.82                     & 14.59                     \\
                                & \quad +HAO  &  13.54 & 13.19 & 11.89 & 12.54   & 13.49       & 15.54        & \textbf{52.22}       & 18.92                    & 13.49                   \\
                                & \quad +CANA     & \textbf{45.33} & \textbf{45.88} & \textbf{37.73} & \textbf{43.28} & \textbf{46.03} & \textbf{23.44}   & 34.78            & \textbf{39.49 (  $\uparrow$ 166.47)} & \textbf{43.28 (  $\uparrow$ 196.58)} \\
                                 \cmidrule(lr){2-11}
                                & TDGIA     & 10.94          & 20.24          & 20.04          & 15.49          & 20.24          & \textbf{20.09}   & 26.34            & 19.06                     & 20.09                     \\
                                & \quad +HAO      & 8.15           & 9.45           & 8.15           & 8.15           & 9.45           & 15.79            & \textbf{50.47}   & 15.66                     & 9.45                      \\
                                & \quad +CANA     & \textbf{31.73} & \textbf{33.48} & \textbf{24.29} & \textbf{28.79} & \textbf{33.73} & 16.89            & 25.84            & \textbf{27.82 (  $\uparrow$ 46.01)}  & \textbf{28.79 (  $\uparrow$ 43.28)}  \\
                                \cmidrule(lr){2-11}
                                & G-NIA     & 8.15           & 8.15           & 8.15           & 8.15           & 8.15           & 22.99            & 17.99            & 11.67                     & 8.15                      \\
                                & \quad +HAO      & 8.15           & 8.60           & 8.15           & 8.20           & 8.60           & 11.19            & 14.79            & 9.67                     & 8.60                      \\
                                & \quad +CANA     & \textbf{45.83} & \textbf{50.57} & \textbf{45.18} & \textbf{49.03} & \textbf{50.42} & \textbf{36.38}   & \textbf{24.04}   & \textbf{43.06 (  $\uparrow$ 268.93)} & \textbf{45.83 (  $\uparrow$ 462.58)} \\
                                                           
\midrule
\multirow{10}{*}{\tabincell{c}{ogbn-\\arxiv}}    & Clean     & 30.01          & 30.33          & 29.86          & 30.28          & 30.32          & 37.32            & 27.66            & 30.82                     & 30.28                     \\
\cmidrule(lr){2-11}
                                & PGD       & 28.61          & 33.00          & 28.99          & 29.64          & 32.66          & \textbf{54.46}   & 31.64            & 34.14                     & 31.64                     \\
                                & \quad +HAO      & 51.98          & 53.16          & 51.71          & 52.60          & 53.19          & 52.54            & 40.68            & 50.84                     & 52.54                     \\
                                & \quad +CANA     & \textbf{52.13} & \textbf{56.02} & \textbf{52.68} & \textbf{54.44} & \textbf{55.98} & 54.16            & \textbf{50.57}   & \textbf{53.71 (  $\uparrow$ 57.31)}  & \textbf{54.16 (  $\uparrow$ 71.19)} \\
                                \cmidrule(lr){2-11}
                                & TDGIA     & 28.50          & 29.86          & 28.73          & 28.49          & 29.82          & 51.82            & 32.37            & 32.80                     & 29.82                     \\
                                & \quad +HAO      & 28.52          & 30.15          & 28.83          & 28.49          & 30.14          & \textbf{51.99}   & \textbf{36.04}   & 33.45                     & 30.14                     \\
                                & \quad +CANA     & \textbf{31.83} & \textbf{41.47} & \textbf{34.44} & \textbf{41.14} & \textbf{40.85} & 51.25            & 30.20            & \textbf{38.74 (  $\uparrow$ 18.12)}  & \textbf{40.85 (  $\uparrow$ 36.98)}  \\
                                
								\cmidrule(lr){2-11}
                                & G-NIA     & 28.49          & 30.38          & 28.49          & 28.50          & 30.33          & \textbf{52.94}   & 28.43            & 32.51                     & 28.50                     \\
                                & \quad +HAO      & 32.31          & 34.60          & 33.32          & 34.00          & 34.46          & 38.25            & 31.45            & 34.06                     & 34.00                     \\
                                & \quad +CANA     & \textbf{32.92} & \textbf{34.93} & \textbf{33.74} & \textbf{35.04} & \textbf{34.74} & 44.16            & \textbf{32.51}   & \textbf{35.43 ( $\uparrow$ 9.00)}   & \textbf{34.74 (  $\uparrow$ 21.90)}  \\
\bottomrule
\end{tabular}}
\end{table*}

Evaluating the similarity between distributions can indeed be a complex endeavor. Directly computing the JS divergence (Eq.~\ref{eq:JS}) presents a challenge, as the probability densities of the ego network distributions for injected and original nodes are unknown. 
The loss of the optimal discriminator closely approximates JS divergence, but obtaining the optimal discriminator is challenging. 
Fortunately, we find the representations of ego networks of original normal nodes follow multidimensional Gaussians, and we assume that the imperceptible injected nodes follow multidimensional Gaussians. 
Inspired by Fr\'echet Inception Distance (FID)~\cite{Heus2017FID}\footnote{In previous work, deep representations are proved to capture high-level information of images~\cite{ZhangIESW18}. In our case on the graph, the deep representations of ego networks can model and associate the attributes and structures of nodes.}, 
we call Fr\'echet Distance to compute the distance between two Gaussian distributions by mean and covariance.
We propose Graph Fr\'echet Distance (\textbf{GraphFD}) to measure the distance between distributions of ego networks of injected and normal nodes, evaluating the camouflage of attack methods.
GraphFD is formalized as:
\begin{equation}
\begin{aligned}
\text{GraphFD}
% &=d\left((\boldsymbol{\mu}_{r}, \boldsymbol{\Sigma}_r),\left(\boldsymbol{\mu}_{m}, \boldsymbol{\Sigma}_{m}\right)\right)\\
% &=\left\|\boldsymbol{\mu}_{r}-\boldsymbol{\mu}_{m}\right\|_{2}^{2}+\operatorname{Tr}\left(\boldsymbol{\Sigma}_{r}+\boldsymbol{\Sigma}_{m}-2\left(\boldsymbol{\Sigma}_{r} \boldsymbol{\Sigma}_{m}\right)^{1 / 2}\right),\end{aligned}
=\left\|\boldsymbol{\mu}_{r}-\boldsymbol{\mu}_{m}\right\|_{2}^{2}+\operatorname{Tr}\left(\boldsymbol{\Sigma}_{r}+\boldsymbol{\Sigma}_{m}-2\left(\boldsymbol{\Sigma}_{r} \boldsymbol{\Sigma}_{m}\right)^{1 / 2}\right),\end{aligned}
\end{equation}
where $\boldsymbol{\mu}_{r}$ and $\boldsymbol{\Sigma}_{r}$ are the mean vector and variance matrix of the original normal nodes, $\boldsymbol{\mu}_{m}$ and $\boldsymbol{\Sigma}_{m}$ are those of the injected nodes. 
% $\operatorname{Tr}$ denotes the trace of a matrix.

\textbf{Rationality of GraphFD.}
Our GraphFD measures the camouflage of injected nodes under the guide of  distribution similarity.
Such an evaluation metric is general, characterizing the network structure and node attributes at the same time. 
We demonstrate the validity of this metric through two aspects: the consistency between GraphFD and intuitive perceptibility, and the consistency between GraphFD and the accuracy of the detection method, as depicted in Figure~\ref{fig:gfd}. 
%As shown in Figure~\ref{subfig:gfd_percep}, the easier the injected nodes are noticed, the larger the GraphFD. GraphFD reflects the perceptible level very well by monotonically increasing.
Figure~\ref{subfig:gfd_percep} illustrates a monotonic increase in GraphFD as the perceptibility of injected nodes increases. Specifically, as the injected nodes become more noticeable, the GraphFD also becomes larger. This finding suggests that GraphFD is a reliable metric of the perceptible level of injected nodes on a graph.
Figure~\ref{subfig:gfd_detect} illustrates a positive correlation between GraphFD and the accuracy of a detection method for injected nodes. 
The larger the GraphFD, indicating a greater dissimilarity between the distributions, the easier it is to detect the injected nodes.
These findings further support the rationality of using GraphFD as a measure of similarity/dissimilarity between distributions on graphs.

\begin{table*}[]
\caption{Camouflage on various metrics on three datasets. Smaller is better, and the smallest is bolded.}
\label{tab:cam}
%\resizebox{18cm}{23mm}{
\begin{tabular}{l ccc ccc ccc}
\toprule
& \multicolumn{3}{c }{ogbn-products}                                                  & \multicolumn{3}{c }{reddit}                             & \multicolumn{3}{c}{ogbn-arxiv}                               \\
\cmidrule(lr){2-4} \cmidrule(lr){5-7} \cmidrule(lr){8-10}
& CAD $\downarrow$         & Smooth $\downarrow$       & GraphFD $\downarrow$          & CAD $\downarrow$          & Smooth $\downarrow$       & GraphFD $\downarrow$          & CAD $\downarrow$          & Smooth $\downarrow$       & GraphFD $\downarrow$          \\
\midrule
Clean & 0.727          & 0.964          & -                & 0.849          & 1.083          & -                         & 0.350          & 0.570 & -                 \\
\midrule
PGD   & 3.206          & 3.698          &  $3.158\times{10}^2$        & 2.823          & 3.081          & $3.839\times{10}^3$        & 1.731          & 1.947          & $2.179\times{10}^3$        \\
\quad +HAO  & 2.517          & 2.835          &  $1.712\times{10}^3$        & 1.615          & 1.875          &  $1.556\times{10}^3$       & \textbf{0.422} & 0.736          &  $ 2.043\times{10}^3$       \\
\quad +CANA & \textbf{2.295} & \textbf{2.810} & \bm{$1.244\times{10}^2$} & \textbf{1.154} & \textbf{1.568} & \bm{$1.414\times{10}^3$} & 0.538 & \textbf{0.735} & \bm{$2.031\times{10}^3$}
\\   
\midrule
TDGIA & 3.524          & 4.316          &  $7.911\times{10}^2$        & 4.581          & 4.875          & $2.387\times{10}^4$         & 3.344          & 3.578          &  $2.582\times{10}^3$        \\
\quad +HAO  & 10.874         & 12.105         &  $7.391\times{10}^3$         & 10.538         & 11.181         &  $5.957\times{10}^4$        & 3.059          & 3.280          & $2.447\times{10}^3$         \\
\quad +CANA & \textbf{1.495} & \textbf{2.489} & \textbf{61.540}    & \textbf{1.322} & \textbf{1.698} & \bm{$1.491\times{10}^3$} & \textbf{0.745} & \textbf{1.097} & \bm{$2.086\times{10}^3$} \\
\midrule
G-NIA & 23.442         & 24.357         & $2.675\times{10}^4$     & 17.052         & 17.364         & $1.406\times{10}^5$ & 3.889          & 4.157 & $2.218\times{10}^3$          \\
\quad +HAO  & 9.630          & 10.508         &  $5.067\times{10}^3$       & 8.794          & 9.408          & $4.413\times{10}^4$                & 3.406          & 3.813          &  $1.982\times{10}^3$       \\
\quad +CANA & \textbf{1.455} & \textbf{2.062} & \textbf{99.688}  & \textbf{1.094} & \textbf{1.573} & \bm{$898.593$} & \textbf{0.547} & \textbf{0.769} & \bm{$1.524\times{10}^3$} \\
\bottomrule 
\end{tabular}
%}
\end{table*}

\subsection{Attack Performance}
To evaluate the effectiveness and applicability of CANA, We equip three representative attack methods (G-NIA, TDGIA, PGD) with CANA,  as well as compare CANA with the above basic attacks and a SOTA imperceptible attack method HAO.
Experiments are conducted on three benchmark datasets.
%The attack performance on vanilla GNN is in Appendix~\ref{apd:vanilla}.
To better demonstrate the attack ability in practice, we evaluate attacks under representative detection methods (COPOD, PCA, HBOS, IForest, and AutoEncoder), and defense methods (Adversarial training and GNNGuard). Specifically, the input of detection methods is the representation of ego networks capturing both network structure and node attributes. 

%We adopt anomaly detection methods to evaluate the imperceptibility and attack performance of node injection attacks. Because during detection, detectors filter out the perceptible injected nodes, and during GNN classification, nodes with high attack performance cause more damage. The misclassification rate indicates the combined effect of both imperceptibility and attack performance. Specifically, the input of detection methods is the representation of the ego network capturing both network structure and node attributes. The representation is modeled by a pre-trained representation model.

Table~\ref{tab:atk} illustrates the misclassification rate on target nodes after filtering by detection methods or defending by defense methods. 
The misclassification on clean graphs (Clean) is also included here as a lower bound.
We list the average and median of the misclassification rate across all the detection and defense methods (\textbf{Average} and \textbf{Median}) to show the overall attack performance. 
%Considering the average value may be affected by outliers, we also report the median value, i.e., \textbf{Median}.
And we report in parentheses the relative increases of attack performance on Average and Median compared with basic attacks for clear comparison.

The basic attacks, namely PGD, TDGIA, and G-NIA, demonstrate poor performance under detection/defense methods across all datasets.
%TDGIA makes more misclassification on defense methods. Unfortunately, the overall performance (Average and Median) of basic attacks on detection and defense methods are still weak compared to Clean. 
As for the heuristic constraint, although the incorporation of HAO has resulted in performance improvements for certain attacks on certain datasets, such as the Average performance of PGD+HAO on reddit, HAO cannot guarantee consistent improvements for all attacks. For example, the Average performance of TDGIA+HAO and G-NIA+HAO even drops slightly on reddit.
%The misclassification rate of G-NIA+HAO improves slightly when attacking detection methods compared with G-NIA. 
%For TDGIA and PGD, the addition of HAO even degrades the performance. We observe similar results when HAO attacks defense methods. 
The results show that previous attacks are too perceptible to detection/defense methods, making the attacks ineffective in the real world.

The integration of our CANA framework brings significant improvement for all node injection attacks and shows superiority on Average and Median on all datasets.
Our CANA achieves the best attack performance under all detection methods on all datasets.
When defense methods are applied, CANA either outperforms or achieves comparable results to basic attacks and HAO, as some defense methods are not strong enough to distinguish and defend against injected nodes.
Specifically, on Median, CANA achieves a boosting improvement of 196\% for PGD, 43\% for TDGIA, and 462\% for G-NIA on reddit. 
The results demonstrate the effectiveness of our CANA framework.

\begin{figure}
\centering
\subfigure[Clean graph]{
\includegraphics[width=3.2cm]{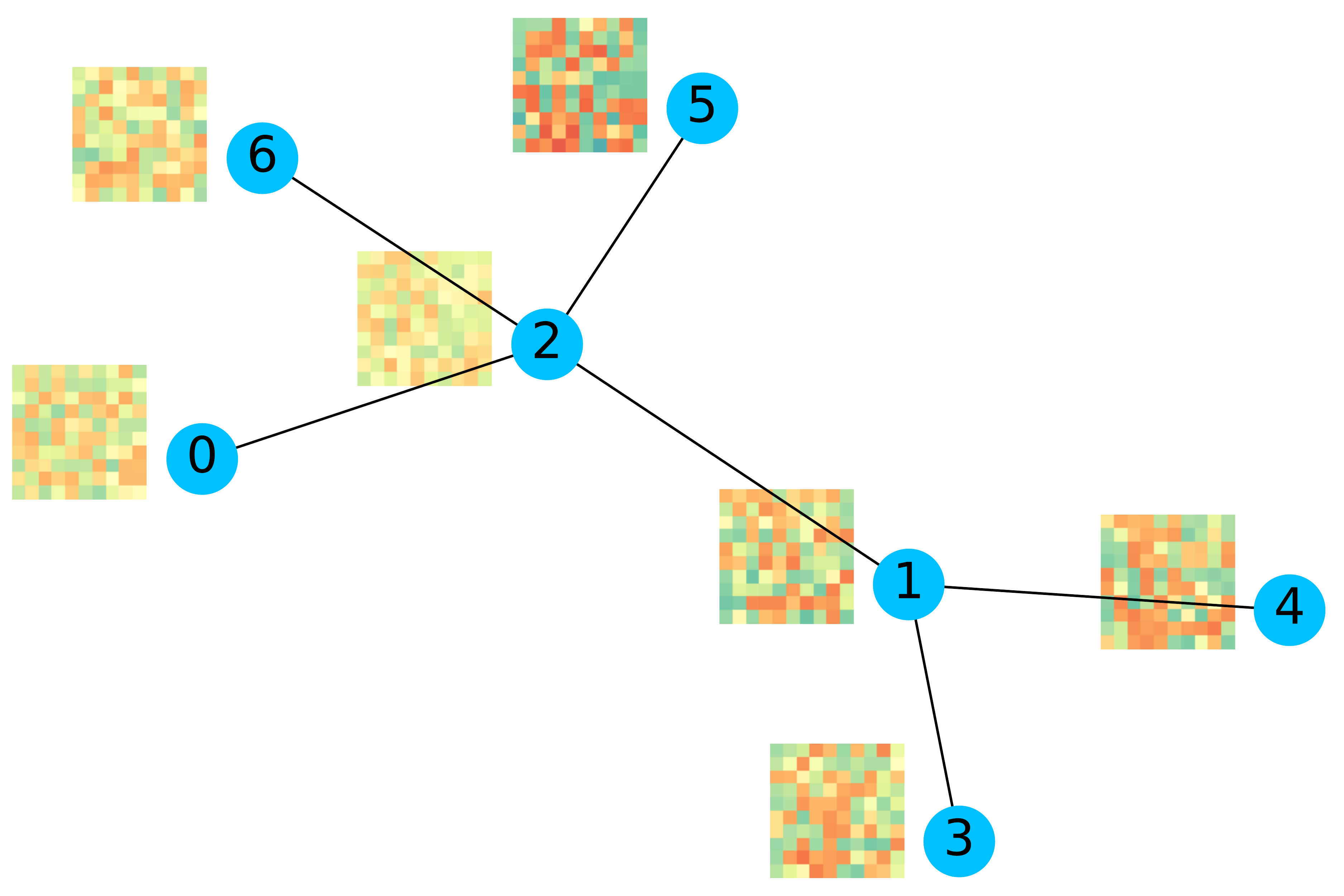}
\label{subfig:case_clean}
}
\quad \quad
\subfigure[G-NIA]{
\includegraphics[width=3.2cm]{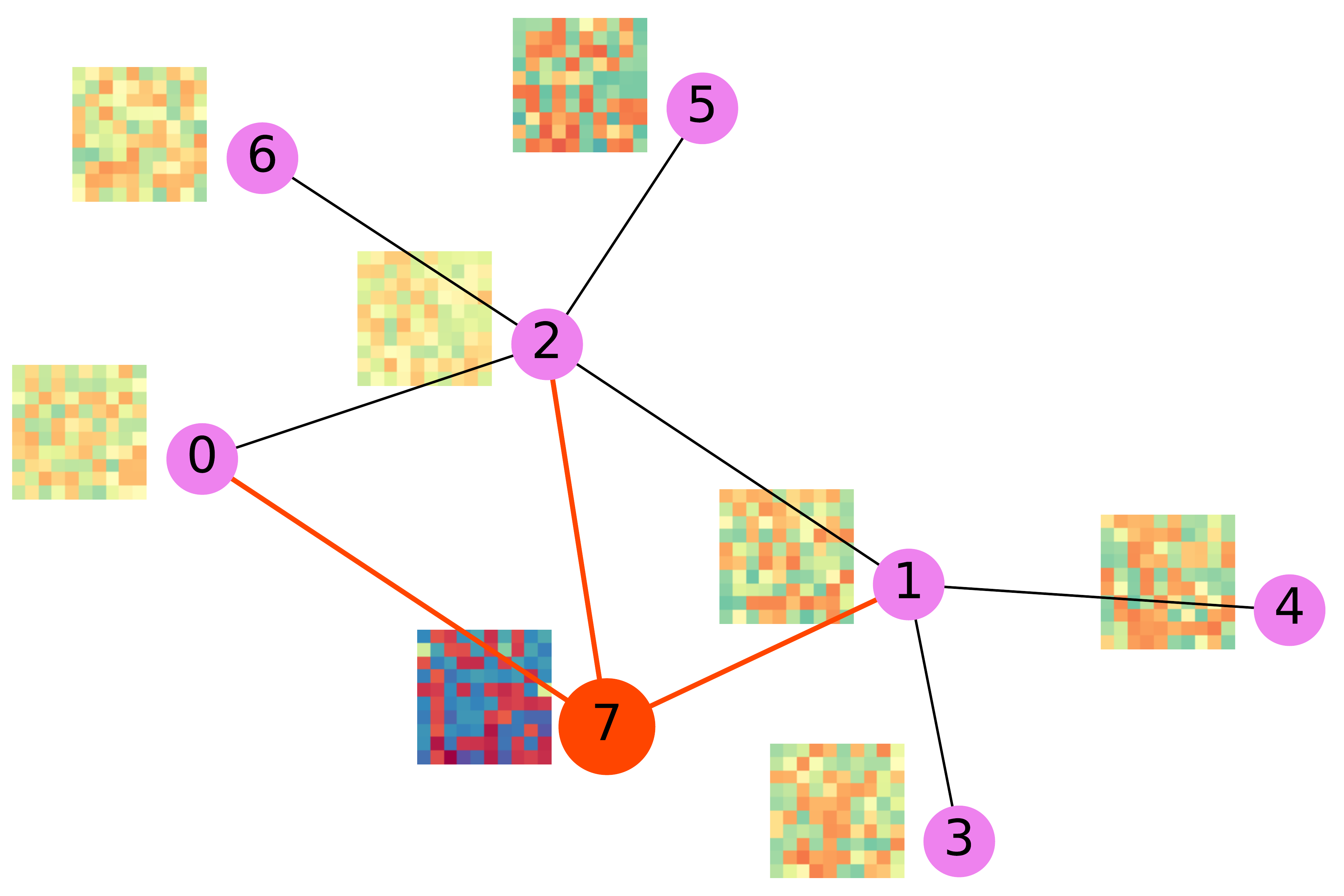}
\label{subfig:case_gnia}
%\end{minipage}%
}
\subfigure[G-NIA+HAO]{
\includegraphics[width=3.2cm]{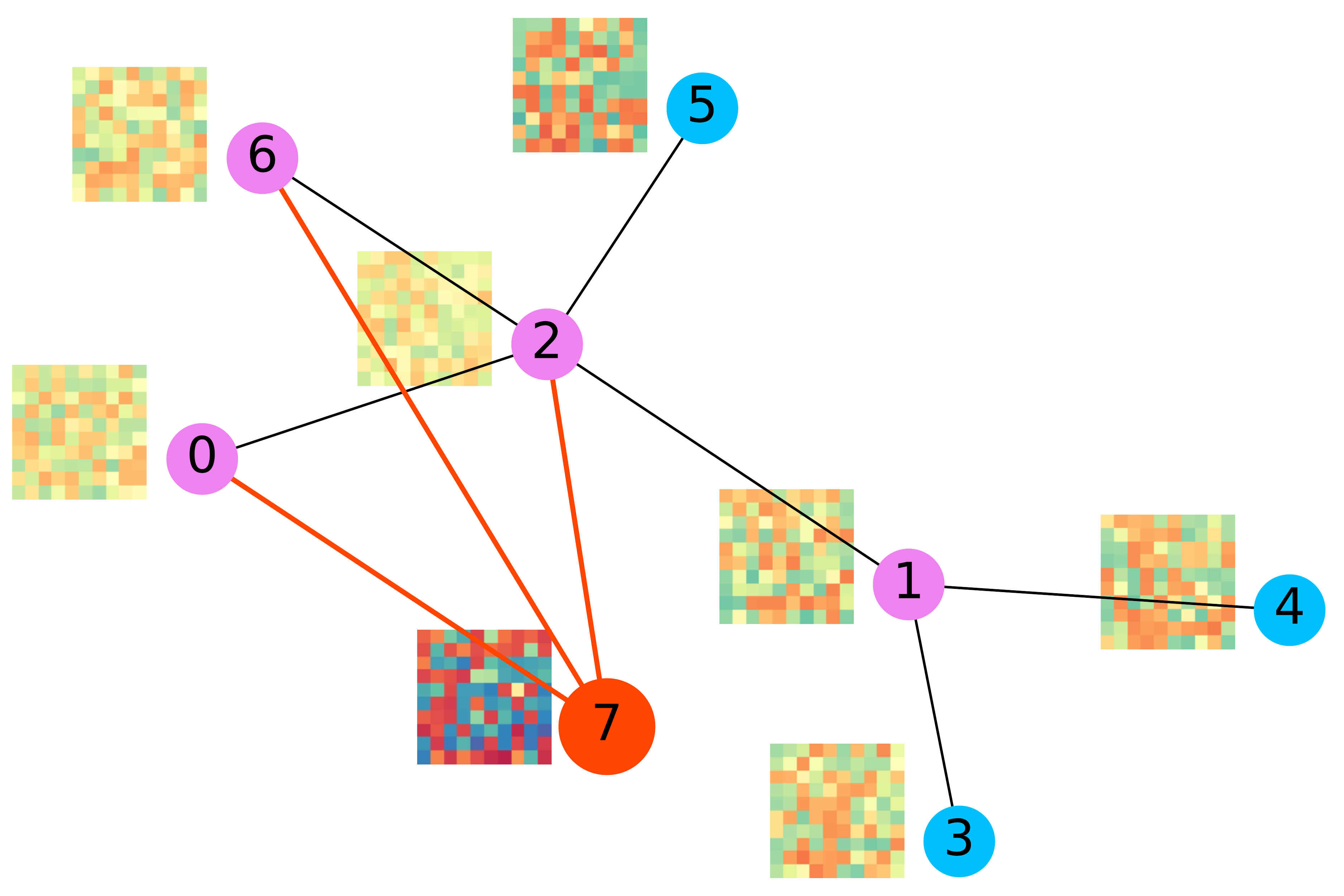}
\label{subfig:case_hao}
}
\quad \quad
\subfigure[G-NIA+CANA]{
\includegraphics[width=3.2cm]{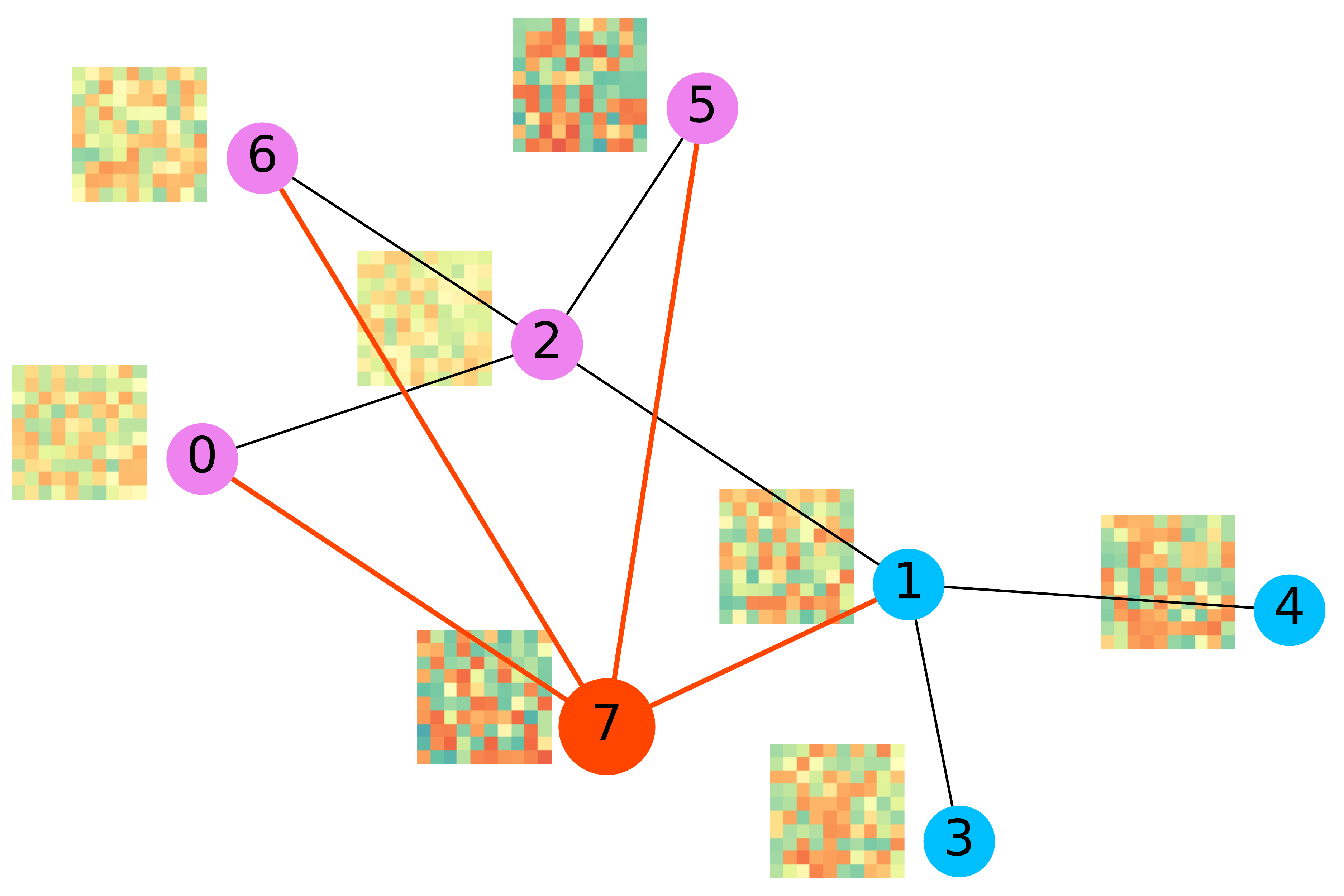}
\label{subfig:case_CANA}
}
\subfigure{
\includegraphics[width=3.6cm]{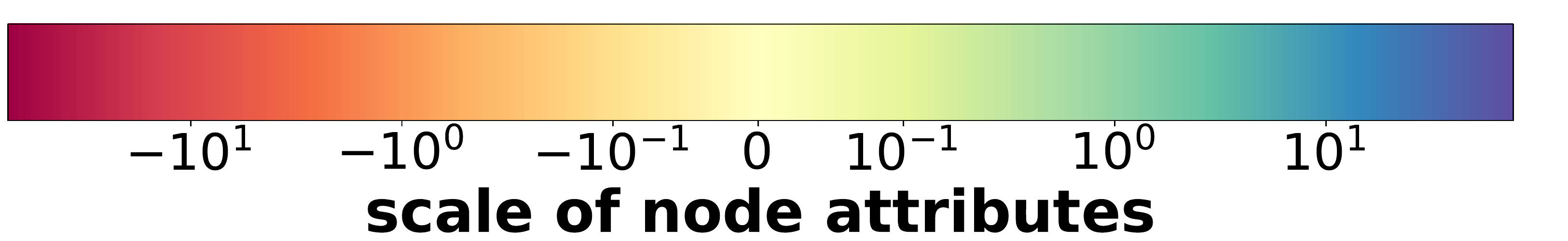}
\label{subfig:case_colorbar}
}
\quad 
\subfigure{
\includegraphics[width=3.3cm]{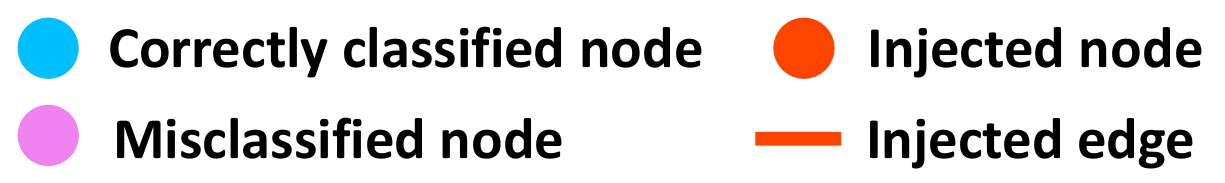}
\label{subfig:case_colorbar}
}
\caption{Visualization of a local network in clean graph and perturbed graphs.
The heatmap represents the node attribute.}
\label{fig:case}
%\vspace{-5pt}
\end{figure}

\subsection{Effect of Camouflage}

To evaluate the camouflage of CANA, we adopt three metrics, i.e., Closest Attribute Distance (\textbf{CAD})~\cite{ZouTDGIA}, Smoothness (\textbf{Smooth})~\cite{chen2022understanding}, and our proposed  GraphdFD.
%Specifically, CAD is the average Euclidean distance between the attribute and its most similar attribute of original normal nodes, i.e., $\text{CAD}_i = \min_{j\in V} d(X_{i}, X_j)$.
%Smooth is the average of node attribute distance to each neighbor, i.e., $\text{Smooth}_i = \frac{1}{|N_i|} \sum_{j\in N_i} d(X_i, X_j)$.
% where $ r_i = \frac{1}{|N_i|} \sum_{j\in N_i} X_j$, $N_i$ means neighbors of $i$.
%The intuition is if there is an original node similar to the injected node, then the injected node is well-camouflaged and imperceptible.
 We report the results of three basic attack methods and imperceptible attack HAO and our proposed CANA on three benchmark datasets.
 The imperceptibility of clean graphs, i.e., Clean, is reported as the upper bound. CAD and Smooth of clean graphs are computed across original nodes, while GraphFD of clean graphs is '-', since there is no injected node in Clean.

As illustrated in Table \ref{tab:cam}, the camouflage of all the basic attacks on all datasets is quite poor. 
%PGD and TDGIA perform better but are still very different from the clean graph.
Although HAO improves the imperceptibility of the specific attack(G-NIA+HAO on ogbn-products), it even decreases the imperceptibility for other attacks(TDGIA+HAO on ogbn-products and reddit).
%Because HAO is heuristic and tries to be imperceptible by optimizing smoothness.  HAO lacks the consideration of the distribution between injected nodes and normal nodes, so it is difficult to narrow the distribution distance such as GraphFD.
This is due to its heuristic nature, which prioritizes smoothness. Additionally, it does not consider the distribution between injected and normal nodes, making it challenging to narrow the distribution distance like GraphFD.
The incorporation of CANA significantly improves the imperceptibility of all basic attacks across all datasets, as evidenced by all metrics.
CAD and Smooth of graphs perturbed by G-NIA+CANA, TDGIA+CANA, and PGD+CANA are very close to the original clean graphs on reddit and ogbn-arxiv.
Since the GraphFD metric cannot be computed on a clean graph, any analysis that takes it into account should compare CANA with HAO and basic attacks
GraphFD of the graphs attacked by CANA shows a significant narrowing when compared to the basic attacks and HAO, indicating that the representation distributions of injected nodes are closer to those of normal nodes. 
Specifically, for G-NIA+CANA, GraphFD decreases by $99.79\%$ on ogbn-products, $19.02\%$ on ogbn-arxiv, and $64.74\%$ on reddit.
The results prove the imperceptibility and camouflage ability of our proposed CANA.

\begin{figure}[t]
\centering
%\vspace{-3pt}
	\includegraphics[width=8.5cm]{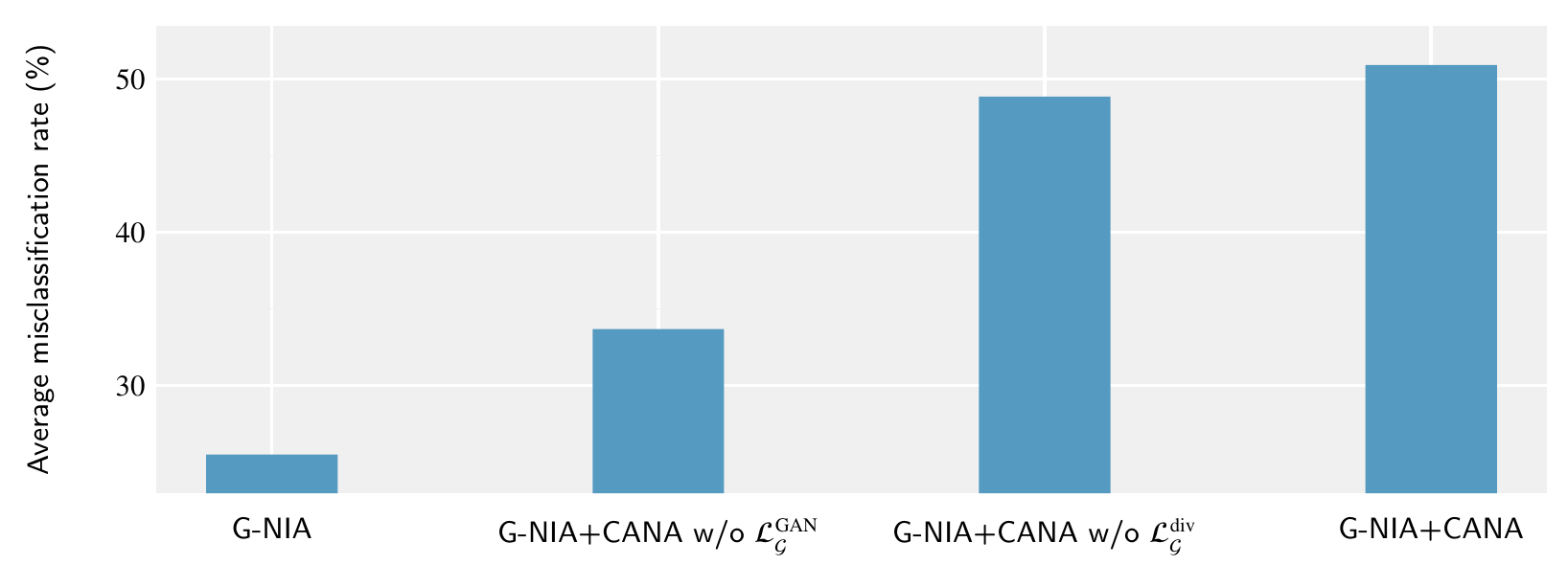}
	\caption{Ablation study.
	Average misclassification rate across detection and defense methods attacked by variants of CANA.}
	\label{fig:abl}
% 	\vspace{-5pt}
\end{figure}

\subsection{Case Study}
To intuitively understand CANA, we take ogbn-products as a case to study the changes before and after node injection attacks.
Figure~\ref{fig:case} visualizes a local network of the clean graph and that of graphs perturbed by G-NIA, G-NIA+HAO, and G-NIA+CANA.
The large red node denotes the injected node. Node color refers to the prediction of a vanilla GCN trained on the clean graph.
Compared with the results on the clean graph, all attack methods successfully cause GNN to classify most nodes incorrectly. 
The basic attack G-NIA injects one node and makes all original nodes misclassified, however, the injected attributes differ significantly from the original attributes, as depicted in the heat map in Figure~\ref{subfig:case_gnia}.
G-NIA+HAO results in the misclassification of four nodes and reduces perceptibility to some extent. Nevertheless, the injected node attributes remain easily detectable.
G-NIA+CANA also makes four nodes misclassified, but the injected node attributes closely resemble those of the normal nodes, demonstrating CANA's exceptional camouflage capabilities.

%According to the heat map, the nodes injected by G-NIA and G-NIA+HAO are obviously different from normal nodes in terms of node attributes.
%In contrast, as shown in Figure~\ref{fig:case}(d), the node attribute injected by G-NIA+CANA is close to the original normal nodes, highlighting CANA's ability to achieve superior camouflage.

\begin{figure}
\centering
\subfigure[Adversarial camouflage loss weight $\alpha$]{
\includegraphics[width=0.4\textwidth]{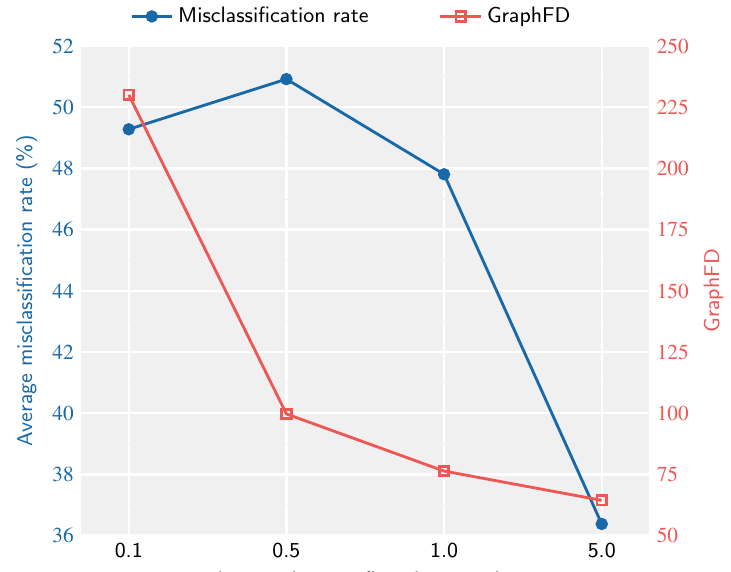}
\label{subfig:case_clean}
}
\quad \quad
\subfigure[Diversity sensitive loss weight  $\beta$]{
\includegraphics[width=0.4\textwidth]{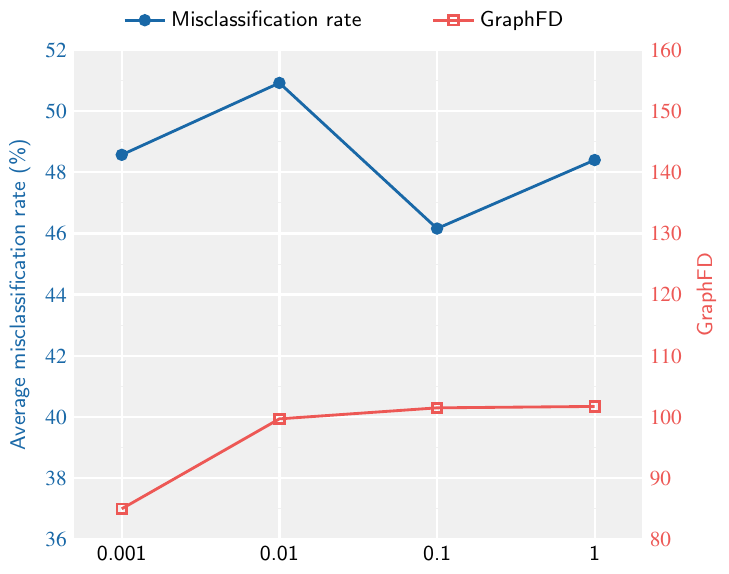}
\label{subfig:case_gnia}
%\end{minipage}%
}
\caption{The influence of $\alpha$ and $\beta$ on attack performance and camouflage on the ogbn-products. The red lines refer to the misclassification rate on the left ordinate, and the blue lines refer to GraphFD on the right ordinate.}
\label{fig:para}
%\vspace{-3pt}
\end{figure}

\subsection{Ablation Study}

We conduct an ablation study to analyze the effectiveness of each part of CANA.
We implement two variants of CANA including CANA without adversarial camouflage loss $\mathcal{L}_{\mathcal{G}}^\text{GAN}$ (CANA w/o $\mathcal{L}_{\mathcal{G}}^\text{GAN}$), and CANA without diversity sensitive loss $\mathcal{L}_{\mathcal{G}}^\text{div}$ (CANA w/o $\mathcal{L}_{\mathcal{G}}^\text{div}$),  where we set  $\alpha=0$ and $\beta=0$, respectively. 
We take the results on ogbn-products with G-NIA as an example, as the results on other datasets with other injection attack methods are similar.
%The experiments are conducted on ogbn-products dataset with the basic attack method G-NIA.

Figure~\ref{fig:abl} presents the average attack performance across all detection and defense methods.
The misclassification rates of CANA without $\mathcal{L}_{\mathcal{G}}^\text{GAN}$ and CANA without $\mathcal{L}_{\mathcal{G}}^\text{div}$ are significantly lower than those of the complete CANA, emphasizing the importance of our losses.
Both variants can unilaterally increase the rate of misclassification under the defense/detection methods compared to the basic attack G-NIA.
Specifically, CANA without $\mathcal{L}_{\mathcal{G}}^\text{GAN}$  causes a misclassification rate of $48.87\%$, and CANA without $\mathcal{L}_{\mathcal{G}}^\text{div}$ yields a misclassification of  $33.70\%$. 
The results also indicate that adversarial camouflage loss is more important than diversity sensitive loss, because the former helps improve the camouflage of injected nodes by using the discriminator as a guide.
Overall, each loss of our CANA plays a crucial role, and these results demonstrate its significance in attack performance under detection/defense methods.

\subsection{Hyper-parameter Analysis}
\label{apd:hyper_ana}
We analyzed the sensitivity of our CANA to the hyper-parameters $\alpha$ and $\beta$, from two perspectives: attack performance and node camouflage.
The attack performance refers to the average misclassification rate across all detection and defense methods, and the camouflage refers to GraphFD.
% are illustrated in Figure \ref{fig:abl}.
%We analyze the sensitivity of $\alpha$ and $\beta$ on the camouflage and attack performance on ogbn-products.
%The results are illustrated in Figure \ref{fig:abl}.
%To demonstrate the effectiveness of $\mathcal{L}_{\mathcal{G}}^\text{GAN}$ on the camouflage metric, we evaluate the attack performance and GraphFD under different weights of $\alpha$. 
We vary the $\alpha$ in $\{0.1,0.5,1.0,5.0\}$, $\beta$ in $\{10^{-3}, 10^{-2}, 10^{-1}, 1\}$. 
As shown in Figure \ref{fig:para} (a), GraphFD (node camouflage) can be consistently improved with the increases of $\alpha$, which validates the effectiveness of $\mathcal{L}_{\mathcal{G}}^\text{GAN}$ on improving the camouflage of injected nodes. 
The attack performance initially improves with smaller values of $\alpha$ but degrades as $\alpha$ increases.
Specifically, when $\alpha$ is large ($\alpha=5.0$), the dominance of camouflage in the attack task inevitably leads to a reduced attack success rate. 
The optimal trade-off between attack strength and node camouflage is achieved at $\alpha=0.5$, resulting in the best attack performance.
As depicted in Figure~\ref{fig:abl} (b), the misclassification rate and GraphFD are observed to be insensitive to changes in $\beta$, as $\mathcal{L}_{\mathcal{G}}^\text{div}$ is primarily used to prevent mode collapse, rather than to directly enhance the camouflage.

\textbf{Further Discussions.}
We find that there is a trade-off between the attack performance and node camouflage under certain defense/detection mechanisms. 
This is because camouflage attacks can effectively evade detection/defense through appropriate node camouflage, an excessively camouflaged attack may compromise performance due to its high similarity to clean data. 
In the absence of any detection/defense mechanism, non-camouflage attacks exhibit better attack performance, as exemplified by G-NIA achieving a misclassification rate of up to 99\%.  
However, in practical scenarios, the presence or absence of a detection/defense mechanism is often unclear, leading us to contemplate whether to employ a camouflaged attack.
One possible way is first injecting a small number of nodes, and then evaluating the attack success rate of these injected nodes to determine whether a camouflaged attack is necessary.
We believe this to be an intriguing research direction and intend to further investigate it in the future.

\begin{table}[]
\caption{Training time comparison of node injection attacks.}
\label{tab:time}
\begin{tabular}{l|ccc|ccc|ccc}
\toprule
 &PGD   & +HAO  & +CANA  & TDGIA & +HAO & +CANA & G-NIA & +HAO  & +CANA    \\
Time (minute) & 0.22 & 0.20 & 1.79 & 3.02  & 2.91 & 16.21 & 61.06 & 75.53 & 200.65  \\  
\bottomrule    
\end{tabular}
\end{table}

\subsection{Time analysis}
To clarify CANA's computation time, we compare the complete training time of CANA, HAO, and basic attacks (PGD, TDGIA, and G-NIA), using a single NVIDIA V100 32 GB GPU. Taking ogbn-products as an example, we report the results in Table~\ref{tab:time}. The time of CANA is associated with the corresponding basic attacks. Specifically, the training time of PGD+CANA is approximately 8.1 times that of PGD training, TDGIA+CANA is nearly 5.4 times that of TDGIA, and G-NIA+CANA is about 3.3 times that of G-NIA. These results demonstrate that CANA necessitates only minimal additional computation, which will not hinder its practical application on large-scale datasets.

\section{Conclusion and Future Work}

 In this paper, we find that the malicious nodes generated by existing node injection attack methods are easy to be identified by defense methods. 
 To solve the issue, we first formulate the camouflage on graphs as the distribution similarity between the ego networks centering around the injected nodes and  those centering around the normal nodes, characterizing both network structures and node attributes. 
 Then we propose an adversarial camouflage framework for node injection attacks, namely CANA, to improve the camouflage of injected nodes through an adversarial paradigm.
CANA is a general framework, which could be attached to any existing node injection attack methods, improving node camouflage while inheriting the performance of existing node injection attacks.
A novel metric, i.e., GraphFD, is further proposed for evaluating the imperceptibility of injected nodes.
Extensive experimental results demonstrate that equipped with our CANA framework, existing node injection attack methods significantly improve the attack performances under various detection and defense methods with better imperceptibility/camouflage. 

%We believe the camouflage of graph adversarial attack is an interesting new direction. There are some valuable research problems, such as the reason why there is a trade-off between attack performance and camouflage. We are excited about the potential of this research direction and plan to conduct further investigations in the future.

\textbf{Practical implementation.} First, we train a surrogate model like other attacks. Next, we use an attack method to inject nodes. Since the presence or absence of a detection/defense mechanism is often unclear, we start by applying the basic attack method to inject 1\% of the node budget. If the attack fails, it indicates the presence of a defense/detection method in the system. In this case, we integrate CANA to inject the remaining nodes for a camouflaged attack. Otherwise, if the attack is successful, it implies no defense or detection method is present, allowing us to rely solely on the basic attack.

\textbf{Practical implications.}
Node injection attacks can significantly degrade the GNN performance, and the practical implications of these attacks largely depend on GNN applications, such as fraud detection, recommender system, bioinformatics, and so on. In security-critical scenarios, detection/defense methods are often used to distinguish and remove injected nodes. However, our CANA approach renders injected nodes indistinguishable from normal ones, making them imperceptible to defense or detection methods, thereby increasing their threat in practical scenarios. This study highlights the potential risks of node injection attacks when deploying GNNs in real-world settings and emphasizes the need to address the security vulnerabilities of GNNs in practical applications.

\textbf{Future work.}
We view the camouflage of graph adversarial attacks as a fascinating and promising research direction. There are several valuable research questions, such as studying the trade-off between attack performance and camouflage under certain defense/detection mechanisms from both empirical and theoretical perspectives, exploring the more tractable loss functions for addressing camouflage, and experiments on more downstream tasks. We are enthusiastic about exploring directions further and plan to conduct additional investigations in the future.

%This work mainly focuses on the camouflage of node injection attacks. A potential future direction is to extend the CANA framework to other attack scenarios on graphs, such as modifying existing network structures or node features. Further exploration of graph adversarial attacks can raise awareness among researchers about the security vulnerabilities of GNN applications.

\section*{Acknowledgments}

This work is funded by the National Key R\&D Program of China (2022YFB3103700, 2022YFB3103701), and the National Natural Science Foundation of China under Grant Nos. 62102402, U21B2046, 62272125. Huawei Shen is also supported by Beijing Academy of Artificial Intelligence (BAAI).


%\bibliographystyle{model1a-num-names}
\bibliographystyle{cas-model2-names}
%\bibliographystyle{elsarticle-num}

\bibliography{CANA.bib}
%\input{CANA.bbl}

%\clearpage
%%\appendix
%%
%%
%\subsection{Attack Performance on Vanilla GNN}
%\label{apd:vanilla}
%Figure~\ref{tab:vanilla} shows  that the attack performance on vanilla GCN without any defense or detection methods on three datasets.
%
%\begin{table}[]
%\caption{Misclassification rate (\%) of node injection attacks on GCN. Larger is better, and the largest one is bolded}
%\label{tab:vanilla}
%\begin{tabular}{c|c|c|c}
%\toprule
%      & ogbn-products & reddit             & ogbn-arxiv         \\
%\midrule
%Clean & 21.01                         & 8.15            & 28.49          \\
%\midrule
%G-NIA  & \textbf{99.95}                & 99.85           & \textbf{98.81} \\
%+HAO  & 99.81                         & \textbf{100.00} & 98.48          \\
%+CANA & 76.66                         & 63.82           & 45.48          \\
%\midrule
%TDGIA & 93.95                         & \textbf{99.80}  & 79.36          \\
%+HAO  & \textbf{97.00}                & 97.25           & \textbf{80.57} \\
%+CANA & 52.74                         & 42.63           & 49.02          \\
%\midrule
%PGD   & \textbf{97.52}                & \textbf{99.80}  & \textbf{99.93} \\
%+HAO  & 91.38                         & 94.80           & 93.84          \\
%+CANA & 90.52                         & 57.92           & 53.12   \\
%\bottomrule      
%\end{tabular}
%\end{table}

% \subsection{Performance on clean graphs}

% \section{Visualizations}

%% Biography
%% Here goes the biography details.

%% Here goes the biography details.

\end{document}